\documentclass[review]{elsarticle}

\usepackage{hyperref}
\usepackage{longtable}
\usepackage{amsmath,amsthm,amssymb,epsfig,bm,amsmath}
 \usepackage{graphicx}
\usepackage{float}
\usepackage{amsmath}
\usepackage{tabularray}
\usepackage{mhchem}
\usepackage{xparse}
\usepackage{adjustbox}
\usepackage{MnSymbol}
\usepackage{mathdots}
\usepackage{makecell}
\usepackage{listings}
\usepackage{footnote}
\usepackage[flushleft]{threeparttable}
\usepackage{multirow}
\usepackage{relsize}
\usepackage{lettrine}
\usepackage{courier}
\usepackage{color} 
\definecolor{mygreen}{RGB}{28,172,0} 
\definecolor{mylilas}{RGB}{170,55,241}
\usepackage{amsmath,amsthm,amssymb,amsfonts} 
\usepackage{graphicx} 
\usepackage[margin=1in]{geometry} 
\usepackage[yyyymmdd]{datetime}
\usepackage{algorithm}
\usepackage[noend]{algpseudocode}
\usepackage{lipsum}
\usepackage{setspace}
\usepackage{siunitx}
\usepackage{pdfpages}
\usepackage{tikz}
\usetikzlibrary{positioning}
\usepackage{booktabs,caption}
\usepackage[]{hyperref}
\usepackage{soul}
\usepackage{xcolor}
\usepackage{subcaption}

\hypersetup{
 colorlinks  = true, 
 urlcolor   = blue, 
 linkcolor  = black, 
 citecolor  = blue 
}
\usepackage[numbers]{natbib}
\usepackage{color}
\definecolor{codegreen}{rgb}{0,0.6,0}
\definecolor{codegray}{rgb}{0.5,0.5,0.5}
\definecolor{codepurple}{rgb}{0.58,0,0.82}
\definecolor{backcolour}{rgb}{0.95,0.95,0.92}
\lstdefinestyle{mystyle}{
  backgroundcolor=\color{backcolour},  
  commentstyle=\color{codegreen},
  keywordstyle=\color{magenta},
  numberstyle=\tiny\color{codegray},
  stringstyle=\color{codepurple},
  basicstyle=\footnotesize,
  breakatwhitespace=false,     
  breaklines=true,         
  captionpos=b,          
  keepspaces=true,         
  numbers=left,          
  numbersep=5pt,         
  showspaces=false,        
  showstringspaces=false,
  showtabs=false,         
  tabsize=2,
  escapeinside={<@}{@>},
}
 
\usepackage{booktabs} 
\usepackage{siunitx} 
\usepackage{comment}
\usepackage{mathtools} 
\usepackage{gensymb} 
\usepackage{mhchem} 
\usepackage{fixltx2e} 
\usepackage{bbm}

\usepackage{multirow}

\usepackage{fancyhdr} 
\theoremstyle{definition}

\theoremstyle{definition}

\theoremstyle{remark}

\usepackage{soul} 
\soulregister\cite7
\soulregister\ref7
\soulregister\pageref7
\usepackage{bm}

\usepackage{framed} 

\usepackage{multicol} 

\usepackage{nomencl} 
\usepackage{tikz}
\makenomenclature
\usepackage[resetlabels,labeled]{multibib}

\renewcommand*\nompreamble{\begin{multicols}{2}}
\renewcommand*\nompostamble{\end{multicols}}

\usepackage{amssymb}
\usepackage{pifont}
\definecolor{light-gray}{gray}{0.95}

\def\R{{\mathbb R}}

\usepackage[left, pagewise]{lineno}

\journal{Elsevier}

\begin{document}


\begin{frontmatter}

\title{\large Opening the Black-Box: Symbolic Regression with Kolmogorov-Arnold Networks for Energy Applications}

\author{Nataly R. Panczyk$^{a}$, Omer F. Erdem$^{a}$, Majdi I. Radaideh$^{a,*}$}

\cortext[mycorrespondingauthor]{Corresponding Authors: Nataly Panczyk (npanczyk@umich.edu), Majdi I. Radaideh (radaideh@umich.edu)}

\address{$^{a}$Department of Nuclear Engineering and Radiological Sciences, University of Michigan, Ann Arbor, MI 48109, United States}

\begin{abstract}
\small
While most modern machine learning methods offer speed and accuracy, few promise interpretability or explainability-- two key features necessary for highly sensitive industries, like medicine, finance, and engineering. Using eight datasets representative of one especially sensitive industry, nuclear power, this work compares a traditional feedforward neural network (FNN) to a Kolmogorov-Arnold Network (KAN). We consider not only model performance and accuracy, but also interpretability through model architecture and explainability through a post-hoc SHAP analysis. In terms of accuracy, we find KANs and FNNs comparable across all datasets, when output dimensionality is limited. KANs, which transform into symbolic equations after training, yield perfectly interpretable models while FNNs remain black-boxes. Finally, using the post-hoc explainability results from Kernel SHAP, we find that KANs learn real, physical relations from experimental data, while FNNs simply produce statistically accurate results. Overall, this analysis finds KANs a promising alternative to traditional machine learning methods, particularly in applications requiring both accuracy \textit{and} comprehensibility. 

\end{abstract}

\end{frontmatter}


\setstretch{1.3}

\section{Introduction}
\label{sec:intro}
The boom of artificial intelligence (AI) and machine learning (ML) has peppered industries across the world with opportunities for enhanced efficiency, improved safety, and increased production. Obtaining these features, however, costs users the understanding of how they got from Point ``A'' to Point ``B.'' Engineers refer to this condition as a ``black-box,'' but a layperson would not be far off to label it as \textit{magic}. The harrowing difference between AI and magic, though, is that AI/ML can be wrong. One can accept this kind of blind faith when the stakes are low. But when public safety is on the line, as it often in sensitive industries including finance, medicine, aerospace, or nuclear power, we must demand a more rigorous, transparent approach. Sensitive industries that stand to benefit from AI require ML models that offer both \textit{explainability} and \textit{interpretability}. 

The U.S. Office of Management and Budget outlined a broad demand for this in a March 2024 memorandum stating that federal agencies should ensure sufficient oversight for operators to ``interpret and act'' on an AI's output and to monitor all AI systems for unlawful discrimination and bias, a near-impossible task for status quo black-box algorithms \cite{shalanda_d_young_memorandum_2024}. As a particularly sensitive application, the U.S. Nuclear Regulatory Commission (NRC) highlighted this need in their \textit{2023-2027 Artificial Intelligence Strategic Plan} (NUREG-2261) \cite{us_nuclear_regulatory_commission_nureg-2261_nodate}. Further, the Federal Aviation Administration (FAA) explicitly calls for research on ``exploring the explainability of AI systems'' in its 2024 \textit{Roadmap for Artificial Intelligence Safety Assurance} \cite{federal_aviation_administration_roadmap_2024}. While regulation demands compliance, we also must acknowledge a generally surging demand for transparent AI systems, because trusting a model precedes trusting its output.

To meet this demand for model interpretability and explainability, we propose a Kolmogorov-Arnold Network (KAN). Most modern machine learning methods rely on the Universal Approximation Theorem (UAT); KANs rely on their namesake, the Kolmogorov-Arnold Theorem (KAT). KAT states that any multivariate continuous function can be represented as a superposition of univariate continuous functions \cite{an_kolmogorov_representation_1957}. Unlike a multi-layer perceptron (MLP) network, which places activation functions on the nodes of its network, a KAN's activation functions are on its edges. MLP will be referred to as Feedforward Neural Network (FNN), which is the popular AI/ML regression option in various engineering and energy applications \cite{radaideh2019combining,talaat2020load,saleem2020application,khosravi2018prediction}. A KAN uses a combination of B-splines to generate these activation functions, which can later transform into symbolic expressions. Though fairly nascent compared to FNN, KANs have already demonstrated tremendous potential; Ji, Hou, and Zhang \cite{ji_comprehensive_2025} and Somvanshi et al. \cite{somvanshi_survey_2024} describe a variety of KAN applications to date.

\subsection{Literature Review and Research Gaps}
Despite a growing demand and interest for KANs, the literature lacks applications of KANs for the sensitive industries, including energy industries, that demand interpretable and explainable AI the most. In our review, we found four key themes of overlap between our work and the existing literature: explainable AI, interpretable AI, applications of KANs compared to traditional FNNs, and comprehensible AI for sensitive applications. In this section, we will discuss how our work aligns at the intersection of these themes, and why this gap must be filled. 

Explainable AI, perhaps more than any other theme we researched, demonstrated the widest breadth across industries, applications, and methods. Likewise, we found explainable AI (sometimes, XAI) one of the most frequently muddled definitions, along with interpretable AI (sometimes, IAI). For this reason, we provide our own definitions of each in Section \ref{sec:theory}. In terms of explainable AI, most of the literature finds motivation from the Defense Advanced Research Projects Agency (DARPA)'s 2019 report outlining the need for explainable AI as ``human-understandable AI systems through the use of effective explanations'' \cite{gunning_darpas_2019}. From the other side of the Atlantic, some studies also note that the European Union's General Data Protection Regulation (GDPR) could mandate providers to ``provide users with explanations of the results of automated decision-making based on their personal data'' \cite{burkart_survey_2021}. China, another global player, also published white papers and guidelines regarding explainability and transparency that largely overlap with the prior definitions by other countries. This study \cite{speith2024explainability} conducted a comparative
analysis of corporate and national AI ethics guidelines in Germany and
China to uncover how transparency and explainability are operationalized
in practice. Both German and Chinese guidelines emphasize transparency and explainability. German guidelines focus on informed self-advocacy and ensuring AI interactions are clear but remain general. In contrast, Chinese guidelines provide more specific instructions on what to explain and to whom, with less emphasis on individual rights.

Suffice it to say that the explainable AI community is still far from either DARPA's, the EU's, or the Chinese's goals, particularly when considering deep learning methods, but a couple of interesting patterns emerge when evaluating the state-of-the-art. First, it becomes clear that SHAP and LIME are among the two most favored post-hoc explainability methods across industries. Second, many papers indicate a trade-off between model explainability and performance, a claim we find deserving of further investigation in this work. 

To demonstrate some of these ideas, the authors of \cite{kim_quantitative_2025} perform a comparison of various AI methods for nuclear power plant accident diagnoses. They evaluate two accident types as case studies for their analysis-- a steam generator tube rupture and an excess steam demand event, using deep neural networks to evaluate each, coupled with SHAP's integrated gradients method for post-hoc explainability. We find this paper particularly relevant to our work because it also considers analysis in the field of nuclear energy a safety-critical scenario that demands explainable AI. The authors of \cite{kim_quantitative_2025}, like in many other explainable AI analyses, default to the notion that a ``rough trade-off exists between the explainability of AI models and their complexity or performance." While this seems reasonable at face value, this statement largely oversimplifies model performance across datasets (see \cite{rudin_stop_2019}) and neglects models with more accessible explainability analyses, like KANs. A related study without explainable AI \cite{radaideh2020neural} also successfully predicted transient nuclear accidents with high accuracy using deep neural networks and long short-term memory. However, this came at the expense of creating a black-box model for a safety-critical scenario. The literature contains similar works on photovoltaic modeling \cite{gomes_leveraging_2025}, wastewater treatment plant modeling \cite{sheik_insights_2025}, building energy management system modeling \cite{gonzalez-briones_evolution_2025}, LDL cholesterol prediction and classification \cite{sezer_explainable_2024}, cybersecurity modeling via intrusion detection systems \cite{al_explainable_2025}, advanced oxidation system modeling \cite{huang_interpretable_2023}, and credit card default predication modeling \cite{talaat_toward_2024}. Each of these analyses seek to extract underlying knowledge from their models, with hopes to gain model trust and improve their understanding of the mechanisms driving their dataset. Instead, they produce statistically-backed feature importances that put little more than a small crack into the black-box of the algorithms they employ. It is worth mentioning that the physical \textit{truth} of those feature importances cannot be necessarily guaranteed. Instead, they represent the extent to which a given feature influences the model’s predictions, with its importance fluctuating depending on the features included in the model, regardless of any physical significance.

Upon switching our search focus to ``interpretable AI,'' we found another two major ideas: one, ``interpretable'' is frequently misused as a synonym for ``explainable'' under our definitions (and those from authors who make distinctions between the two, including \cite{bertsimas_explainable_2023}, \cite{burkart_survey_2021}, \cite{sheik_insights_2025}, \cite{schramm_comprehensible_2023},  \cite{minh_explainable_2022}, and \cite{zhang_explainable_2022}), and two, interpretable AI frequently reduced authors to simplified models, like linear regression and decision trees (i.e., excluded them from neural networks and their derivatives). As previously mentioned, we will present a full definition distinguishing explainable and interpretable AI in Section \ref{sec:theory}, but roughly, \textit{explainable} AI tells you \textit{why} a model produced a certain output and \textit{interpretable} AI tells you \textit{how} it produced that output (\cite{gilpin_explaining_2019}, \cite{hare_explainable_2022}). Several papers (including \cite{huang_interpretable_2023} and \cite{han_application_2025}) we reviewed conducted SHAP analyses on black-box models and deemed them ``interpretable.'' This definition did not necessarily align with our work that clearly distinguishes these two terms, so we categorized these works with other explainability-focused papers.

As a second feature, many works in need of fully interpretable AI resorted to classical machine learning methods at risk of sacrificing accuracy. For example, one work which sought to predict blood glucose level in patients with type 1 diabetes exclusively tested inherently interpretable models, including decision trees and various regressions \cite{basile_blood_2025}. This work demanded transparency to patients; neural networks were simply not feasible. Fortunately, the authors found strong results using a decision tree, but this study highlights model limitations when true interpretability becomes a requirement. Similar works were performed on modeling interpretable schizophrenia diagnoses \cite{shivaprasad_interpretable_2024}, evaluating interpretable cancer risk modeling \cite{bertsimas_explainable_2023}, evaluating the mechanical performance of sustainable concretes \cite{ulucan_modelling_2024}, and novel methods for generating knowledge graphs \cite{schramm_comprehensible_2023}. Overall, truly interpretable works were rarer, and often demanded in highly personal or otherwise sensitive scenarios.  

Next, to specifically evaluate the state of the field considering the model we propose in this research, we shifted our review towards works on KANs. Here, we found many papers highlighting the high performance of KANs with little to no mention of their inherent interpretability capabilities. While this may initially seem like a simple oversight, this fact alone offers promise for the effectiveness of KANs without tradeoffs for interpretability. One very recent work of particular note by \cite{ansar_comparison_2025} compared KANs to multi-layer perceptrons (MLPs) for energy system optimization, a topic well-aligned with the case studies presented in our work. While this paper did compare the performance of KANs to MLPs \textit{and} performed a post-hoc explainability analysis using SHAP on both models, they did not highlight or evaluate the interpretability potential of the KANs via conversion to symbolic equation and how much that conversion could affect KAN's accuracy. Other papers performed similar analyses but failed to evaluate explainability \cite{peng_predictive_2024}, \cite{shuai_physics-informed_2025}, and  \cite{zhong_interpretable_2024}. Others simply studied KANs for their high performance in a variety of industries, including epileptic seizure prediction \cite{herbozo_contreras_kaneeg_2025}, power system management \cite{yin_golden_2025}, smart grid intrusion detection \cite{wu_graph_2025}, stock price forecasting \cite{haryono_permuted_2024}, streamflow forecasting for Central European rivers \cite{granata_advanced_2024}, credit card fraud detection \cite{le_robust_2024}, energy load forecasting \cite{liu_novel_2025}, and simulating fish swimming behaviors \cite{li_simulating_2025}. 

Various domains in the nuclear industry have adopted FNNs, many of which the datasets in this study are derived from. Their applications include nuclear reactor siting analysis \cite{erdem2025multi}, integration with generative AI through generative adversarial networks (GANs) \cite{nabila2025data}, digital twins \cite{lin2021development}, reinforcement learning policies for nuclear fuel optimization \cite{radaideh2021physics}, surrogate modeling to support reinforcement learning in reactor control \cite{radaideh2025multistep}, fault prognosis \cite{khentout2023fault,radaideh2022application}, surrogate models to optimize proton target systems for particle accelerators \cite{radaideh2022model}, nuclear fuel transmutation \cite{bae2020deep}, and spent fuel modeling \cite{khuwaileh2024once}, among many others. Other potential applications for KAN, where traditional FNN or machine learning have not been extensively utilized, include natural language processing for detecting public sentiment, as demonstrated with large language models in \cite{kwon2024sentiment}, as well as criticality safety \cite{price2019advanced}, shielding analysis \cite{husnain2024machine}, accident analysis \cite{chung2021machine,radaideh2018criticality}, and dosimetry \cite{lam2019predicting}. KAN holds the potential to enhance (or even replace) these applications by providing a more interpretable alternative to the black-box limitations of FNNs.

Finally, the fourth and most idealistic theme of the papers we reviewed appeared as more of an objective than an achievement: \textit{comprehensible AI}. This term, articulated by the authors of \cite{schramm_comprehensible_2023}, describes a sort of combination of explainable and interpretable AI, i.e., a demand for knowing both \textit{why} and \textit{how}. 
As an industry-specific demand, the authors of \cite{hatherley_virtues_2022} state, ``Medical AI should be analyzed as a sociotechnical system, the performance of which is as much a function of how people respond to AI as it is of the outputs of the AI.'' Because patient and doctor trust of an AI system is weighed as highly as its accuracy, this industry requires comprehensible AI; it is the only way to establish such trust. Throughout this review, we observe plenty of highly sensitive industries (financial, medical, nuclear, etc.) either making sacrifices in accuracy for interpretability or deeming post-hoc explanations ``good enough.'' Simultaneously, we see massive progress in the development of a new machine learning method, the KAN, to outperform even transformers with attention mechanisms \cite{granata_advanced_2024}, a highly advanced machine learning method. Still, there exists a gap in connecting our most sensitive industries to such machine learning methods, particularly when combining interpretable, high-performing models with post-hoc explainability analysis to compile the most complete, most accurate, and most comprehensible modeling framework the state-of-the-art has to offer. This work seeks to fill that gap. 

\subsection{Novelty}

Building on the previously identified research gaps, the need for a ML model that ensures both interpretability and explainability remains a top priority for sensitive industries. In this study, we specifically consider KANs and benchmark their performance against FNNs. The focus on FNNs stems from their widespread use and, more importantly, their robustness across all problems examined in this work, where other classical ML methods (e.g., support vector machines, random forests, and decision trees) did not achieve comparable accuracy. We then conducted a post-hoc explainability analysis on the KAN and FNN for each dataset to compare their feature importance rankings. The objective of this work is to assess whether KANs can rival traditional FNNs in terms of accuracy for these datasets and whether the trade-offs—or potential advantages—of KANs justify further exploration for additional problems. Driven largely by the authors’ background, this paper highlights KAN performance in a set of problems relevant to nuclear power, a highly sensitive energy sector. However, the datasets analyzed in this work were validated and extensively used in previous research and encompass a multidisciplinary scope, capturing a broad range of physical phenomena, including radiation transport, fluid flow, thermal-hydraulics, materials degradation, system safety, and nuclear physics. The main contributions of this work can be summarized as follows: 

\begin{itemize}
    \item We explore the potential of KANs to achieve comprehensible AI (offering both explainability and interpretability) when benchmarked across a set of engineering and energy-related problems.

    \item We evaluate the potential of symbolic KANs to compete with traditional FNN regressors by comparing their accuracy and explainability (via feature importance ranking). The authors then seek to establish a connection between the symbolic equations generated by KAN and the mathematical models traditionally used to derive the same outputs from a physics-based perspective.

    \item For completeness, we evaluate KAN's performance across a diverse set of regression problems derived from nuclear power. The datasets present a range of challenges, varying in dataset size, linearity and nonlinearity, input and output dimensionality, underlying physical phenomena, and data sources, including analytical, simulated, and experimental data. These diverse combinations enable a comprehensive assessment of KAN's capabilities.

    \item The work presents compelling findings, showing that a well-tuned KAN achieves strong regression performance in its symbolic form, outperforming FNN in terms of interpretability and generally matching its conclusions in terms of explainability, with the exception of some feature exclusion. Although we observed minor accuracy degradation by KAN in certain cases, KAN outperformed FNN in others. More importantly, the KANs generated algebraic equations of varying lengths for all benchmark problems analyzed in this work—a feature precluded by the structure of the FNNs.
\end{itemize}

This paper will begin with Section \ref{sec:theory} describing the key theoretical differences between KANs and FNNs, as well as a full definition of interpretability and explainability and how we can measure these characteristics of machine learning models. In Section \ref{sec:data}, we will describe each of the eight problems considered in this analysis and their associated datasets. Section \ref{sec:methodology} will walk through the KAN model development procedure, including pre-processing, hyperparameter tuning, spline to symbolic conversion, and explainability analysis. Following this, Section \ref{sec:results} will show the model metrics of all eight KAN models and their corresponding FNNs, as well as their explainability analyses. Finally, in Sections \ref{sec:discussion} and \ref{sec:conclusion}, we will discuss our findings and conclude with some ideas for future work and the opportunities for KANs in sensitive industries.

\section{Theory}
\label{sec:theory}

\subsection{Kolmogorov Arnold Networks}
In 1989, George Cybenko proved that, given enough data to reasonably train a network, any continuous function could be approximated arbitrarily well by a neural network with at least one hidden layer and a finite number of weights \cite{cybenko_approximation_1989}. The feedforward neural network (FNN) and all of its derivatives rely on Cybenko's proof, which we know at the Universal Approximation Theorem (UAT). That comprises almost every single AI algorithm based on neural networks known today. The exception being the Kolmogorov Arnold Network (KAN), which instead relies on the 1957 Kolmogorov-Arnold Representation Theorem (KART). The KART states that a multivariate continuous function can be written as a finite sum of continuous univariate functions \cite{an_kolmogorov_representation_1957}. Mathematically, we can observe this statement as:
\begin{equation}
    f(\vec{x}) = f(\vec{x}_1, ...,\vec{x}_n)=\Sigma_{q=0}^{2n+1} \Phi_q ( \,\Sigma_{p=1}^n \phi_{q,p}(\vec{x}_p)) \,
\label{eq:ka-rep}
\end{equation} 
where $\vec{x}_p \in [0,1]$, $\phi_{q,p}(\vec{x}_p) : [0,1] \rightarrow \R$, and $\Phi_q : \R \rightarrow \R$.

A major barrier to realizing the potential of Eq.\eqref{eq:ka-rep} is that the KART does not mandate smoothness from its one-dimensional functions, thus blocking learning using any kind of gradient descent algorithm. This, coupled with limits in computational power left the KART to sit on the mathematical shelf for about 60 years \cite{girosi_representation_1989}, \cite{ji_comprehensive_2025}. In 2024, realizing that the majority of functions \textit{are} smooth, Liu et al. \cite{liu_kan_2024} implemented the KART into its namesake Kolmogorov Arnold Network (KAN). While outliers and edge cases surely exist, Liu et al. \cite{liu_kan_2024} treats them as such-- edge case non-smooth functions that will never be perfectly captured by a KAN. Though an inherent limitation, Liu et. al. \cite{liu_kan_2024} found it a rare enough problem to proceed in spite of. 

In an effort to achieve more flexible, accurate learning, Liu et al. \cite{liu_kan_2024} implement the KAN by going ``wider and deeper,'' i.e., more than $2n+1$ edges and more than $n$ layers. Structurally, while the edges and nodes framework of a KAN sounds like an FNN, they have dramatic differences. One is that the learnable activation functions reside on the \textit{edges} of a KAN, unlike on the fixed activation functions that reside on the nodes of an FNN. Expanding the KART into a KAN with $L$ layers mathematically becomes:
\begin{equation}
    \text{KAN}(x) = (\Phi_{L-1} \circ \Phi_{L-2} \circ \cdots \circ \Phi_1 \circ \Phi_0)(x)
    \label{eq:KAN}
\end{equation} where $\Phi_L$ is a matrix of trainable activation functions $\phi_{j, i}$ where $j=1, \cdots, n_{L+1}$ and $i=1, \cdots, n_L$, $\circ$ is the composition of functions, meaning applying one function to the result of another function. Alternatively, we can express an FNN as:
\begin{equation}
    \text{FNN}(x) = (W_{L-1} \circ \sigma \circ W_{L-2} \circ \sigma \circ \cdots \circ W_1 \circ \sigma \circ W_0)(x)
    \label{eq:FNN}
\end{equation} where $W_L$ represents the transformation matrix for the $L_{th}$ layer (linear weights and biases) and $\sigma$ is the (non-linear) activation function. A visually obvious difference between Eq.\eqref{eq:KAN} and Eq.\eqref{eq:FNN} is that the FNN separates its linear transformations (weights and biases) from its nonlinear transformations (activation functions), whereas the KAN combines all transformations in a single function, $\Phi$. 

The authors of the ``KAN: Kolmogorov-Arnold Networks" \cite{liu_kan_2024} include further details on the mathematical theory behind KANs. This work will implement KANs using the Python package containing KAN architecture created by Liu et. al. \cite{liu_kan_2024} and aptly named \verb|pykan|. 
 
\subsection{Explainability vs. Interpretability}
\label{sec:exp_vs_int}

The literature considering ``explainable'' and ``interpretable'' AI can leave readers believing that those two qualifications are synonyms. While some may disagree on the nuances of either one, for the purposes of this work, we must distinguish them so each term is distinct and useful. 

``Explainability,'' in the context of this work, refers to one's ability to \textit{explain} (i.e., reason) \textit{why} a model came to a given conclusion. For example, performing feature attribution, which gives input parameters a ``score'' for their overall contribution to an output value, is one way of measuring explainability. If feature attribution analysis is possible, then the model contains some level of explainability, even if in post. Another level of explainability, for the case of a feedforward neural network, would be node and layer attribution scores. This allows a developer to extract information about \textit{why} the model predicted the outcomes it did on a more granular level.

``Interpretability,'' in the context of this work, refers to one's ability to \textit{interpret} (i.e., reproduce) \textit{how} a model came to a given conclusion. A perfectly interpretable model means that a human could reproduce the results, from input data to output data, seeing every step of the way. I.e., they know \textit{how} the model works. Some diluted form of interpretability might allow a human to perform a series of transformations on the input data identical to the model of interest, but at some point the human can no longer follow the model's calculations and thus cannot reproduce the final result. A hopefully obvious form of a perfectly interpretable model is an equation-- given values for a set of variables, a human could reproduce the calculation of a computer using that same equation, every time. Similarly, linear regressions and decision trees also exhibit complete interpretability. \textit{Part of the objective of this work is to determine if a perfectly interpretable model, an equation, generated from a KAN can compete with an FNN.} 

When comparing explainability to interpretability, our distinction boils down to this: 
\begin{enumerate}
    \item If the feature helps answer \textit{why}, it increases explainability.
    \item If the features answers \textit{how}, it increases interpretability.
\end{enumerate}

\subsection{SHapley Additive exPlanations (SHAP)}
\label{sec:shap}
Due to the amorphous nature of \textit{explainability}, a variety of methods and metrics exist to quantify this feature of ML algorithms. SHapley Additive exPlanations (SHAP) are one commonly used in the explainable AI space, as mentioned in Section \ref{sec:intro}. SHAP represents a cooperative game-theory-based performance metric for a given feature's (player's) importance to determining a model's outcome (payoff) \cite{salih_perspective_2025}. We can calculate a ``Shapley value'' $\phi_i$ for feature $i$ as:
\begin{equation}
\phi_i = \displaystyle \sum_{S \subseteq \{1,2,...,d\} \backslash \{i\}} \frac{|S|!(d-|S|-1)!}{d!}\left[f(S \cup \{i\}) - f(S)\right]
\label{eq:exact_shap}
\end{equation}
where $d$ is the number of features in the model, $\frac{|S|!(d-|S|-1)!}{d!}$ is the weight for each subset $S$, which accounts for the permutations of the feature set, $f(S)$ represents the model's prediction using only the features in the subset $S$, and $f(S \cup \{i\}) - f(S)$ represents the marginal contribution of player $i$ to the team $S$, i.e., it measures how much additional value feature $i$ brings to the model prediction.

Calculating standard Shapley values rises in cost proportionally to $O(d\times2^d)$ where $d$ is the number of features in a problem. This can quickly escalate beyond a reasonable computational budget for problems with high-dimensional features, thus, we employ an approximate algorithm known as Kernel SHAP. 
\par
Kernel SHAP uses a weighted kernel as well as local linear approximations to modify and streamline Eq.\eqref{eq:exact_shap} \cite{lundberg_unified_2017}. The local linear approximation portion of this modification exists in and of itself as an explainability metric, known as LIME (Local Interpretable Model-agnostic Explanations) \cite{ribeiro_why_2016}. LIME alone, however, often proves insufficient. Kernel SHAP corrects LIME approximations using its namesake weighting kernel that morphs the linear least squares regression of LIME into a function more emulative of classic SHAP. As described in \cite{lundberg_unified_2017}, Kernel SHAP values can be calculated with some modifications to minimizing LIME's objective function. The objective function LIME employs to calculate its $\phi$ values is:
\begin{equation}
\xi=\underset{g \in \mathcal{G}}{\arg \min } L\left(f, g, \pi_{x^{\prime}}\right)+\Omega(g) .
\end{equation}
where $L$ is the loss function and $\Omega$ is a complexity penalty for $g$, where $g$ is the explanation model. For Kernel SHAP, we can determine $L$, $\pi$, and $\Omega$:

\begin{equation}
\begin{aligned}
\Omega(g) & =0, \\
\pi_{x^{\prime}}\left(z^{\prime}\right) & =\frac{(M-1)}{\left(M \text { choose }\left|z^{\prime}\right|\right)\left|z^{\prime}\right|\left(M-\left|z^{\prime}\right|\right)}, \\
L\left(f, g, \pi_{x^{\prime}}\right) & =\sum_{z^{\prime} \in Z}\left[f\left(h_x^{-1}\left(z^{\prime}\right)\right)-g\left(z^{\prime}\right)\right]^2 \pi_{x^{\prime}}\left(z^{\prime}\right),
\end{aligned}\\
\end{equation}
where $\left|z^{\prime}\right|$ is the number of non-zero elements in $z^{\prime}$ and $M$ is the simplified number of input features, according to \cite{lundberg_unified_2017}. Further details on the derivation of Kernel SHAP from Classical SHAP and LIME are provided by \cite{lundberg_unified_2017}. 

\section{Data}
\label{sec:data}

All of the example problems demonstrated in this work employ datasets analyzed extensively with ML methods in \cite{myers_pymaise_2025}. These datasets feature problems ubiquitous to but not exclusive to nuclear engineering, a sensitive industry highlighted throughout this work as demanding of more comprehensible AI. The type of data (analytical, simulated, or experimental) varies between the examples. The complicated nature of nuclear engineering problems often requires computational models in lieu of traditional experiments; ensuring that KANs perform well for both is thus a pertinent feature of this work. All of these benchmark datasets have been validated and utilized in prior research, and will therefore be described only briefly here. We present a summary of all datasets in Table \ref{tab:dataset_summary}.

\begin{table}[!h]
\centering
\caption{Summary of datasets used in this work.}
\label{tab:dataset_summary}
\begin{tblr}{
  width = \linewidth,
  colspec = {Q[342]Q[196]Q[69]Q[87]Q[133]Q[108]},
  vline{2-6} = {-}{},
  hline{1-2,10} = {-}{},
}
Dataset Name                        & Field of Interest~  & Inputs & Outputs & Type         & Reference \\
Critical Heat Flux (CHF)            & Thermal Hydraulics  & 6      & 1       & Experimental &    \cite{le_corre_jean-marie_benchmark_2023}       \\
Materials \& Fuel Performance (FP) & Materials Science   & 13     & 4       & Simulated    &    \cite{radaideh_surrogate_2020}       \\
Light Water Reactor (LWR)           & Nuclear Energy & 9      & 5       & Simulated    &       \cite{myers_pymaise_2025}    \\
Heat Conduction (HEAT)              & Heat Transfer  & 7      & 1       & Analytical    &     \cite{myers_pymaise_2025}      \\
Microreactor               & Nuclear Energy & 8      & 4       & Simulated    &     \cite{price_multiobjective_2022}      \\
Power Control (PC)                & Nuclear Energy & 6      & 22      & Simulated    &     \cite{radaideh_neorl_2023}      \\
Nuclear Safety (NS)                & System Safety & 4      & 4       & Simulated    &    \cite{bauer_results_1993}       \\
Nuclear Cross-Section (NXS)          & Nuclear Physics & 8      & 1       & Exp. \& Sim.   &    \cite{radaideh_shapley_2019}       
\end{tblr}
\end{table}

\subsection{Critical Heat Flux Dataset}
Critical heat flux (CHF), the condition where heat transfer plummets due to an impenetrable vapor layer between a boiling liquid and heated surface, poses a serious safety concern to light water nuclear reactors (LWRs). The CHF dataset comprises over 60 years of experimental data on vertical flow boiling apparatuses collected by the NRC \cite{le_corre_jean-marie_benchmark_2023}. Underpinned by uniformly heated channels meant to simulate flow boiling conditions in a nuclear reactor (light water), these experiments vary pressure (\verb|P|), channel diameter (\verb|D|), heated length (\verb|L|), mass flux (\verb|G|), inlet temperature (\verb|T_in|), and outlet equilibrium quality (\verb|Xe|) as inputs/features to predict a single output, \verb|CHF|. Therefore, this problem features six inputs and one output. 

\subsection{Materials and Fuel Performance Dataset}
Transient scenarios, such as during reactor start-up or shut-down, can cause problematic pellet-cladding mechanical interactions (PCMIs) for oxide fuels in nuclear reactors. Based on BISON simulations, a structural mechanics computer code to simulate nuclear materials performance, Radaideh and Kozlowski \cite{radaideh_surrogate_2020} generated 400 data points of PCMIs for a 10-pellet PWR fuel rod subject to an average linear heat rate of 40 $\frac{kW}{m}$ (a ramp-up transient scenario). The dataset features 13 inputs based on fuel geometry and physical properties and four outputs. The inputs include:  fuel density (\verb|fuel_dens|), porosity (\verb|porosity|), cladding thickness (\verb|clad_thick|), pellet outer diameter (\verb|pellet_OD|), pellet height (\verb|pellet_h|), gap thickness (\verb|gap_thick|), inlet temperature (\verb|inlet_T|), uranium-235 enrichment (\verb|enrich|), fuel roughness (\verb|rough_fuel|), cladding roughness (\verb|rough_clad|), axial power (\verb|ax_pow|), cladding surface temperature (\verb|clad_T|), and pressure (\verb|pressure|). The outputs include: fission gas production (\verb|fission_gas|), maximum fuel surface temperature (\verb|max_fuel_surface_T|), maximum cladding temperature (\verb|max_fuel_cl_T|), and radial cladding diameter displacement (\verb|radial_clad_T|).

\subsection{Light Water Reactor Dataset}
This simulated dataset, originally presented by Myers et al. \cite{myers_pymaise_2025}, demonstrates neutronics fluctuations and corresponding power level changes in a boiling water reactor (BWR) micro core. Myers et al. \cite{myers_pymaise_2025} generated this dataset using radial cross section information from CASMO-4, which they processed through CMSLINK, and interpreted through SIMULATE-3. The study \cite{myers_pymaise_2025} implemented and modeled the core geometry and physics using SIMULATE-3 as well to generate core attributes related to the performance and safety of the system. The dataset contains nine inputs, including: power shaping zone region (\verb|PSZ|), dominant zone region (\verb|DOM|), vanishing zone A region (\verb|vanA|), vanishing zone B region (\verb|vanB|), core inlet subcooling (\verb|subcool|), control rod group position (\verb|CRD|), coolant mass flux (\verb|flow_rate|), power density (\verb|power_density|), and narrow water gap width ratio (\verb|VFNGAP|). The dataset also contains five outputs that include: effective neutron multiplication factor (\verb|K-eff|), maximum planar-averaged pin power peaking factor (\verb|Max3Pin|), maximum global pin-power peaking factor (\verb|Max4Pin|), enthalpy rise hot channel factor (\verb|F-delta-H|), and the maximum \textit{radial} pin-power peaking factor (\verb|MaxFxy|). 

\subsection{Heat Conduction Dataset} 
This dataset, presented by Myers et al.  \cite{myers_pymaise_2025}, differs in that its data come from an simple numerical solution to a heat conduction problem using 1.5D conduction methods to predicting nuclear fuel centerline temperature. The inputs to this dataset include seven physical parameters: linear heat generation rate (\verb|qprime|), mass flow rate (\verb|mdot|), temperature of the fuel boundary (\verb|Tin|), fuel radius (\verb|R|), fuel length (\verb|L|), heat capacity (\verb|Cp|), and thermal conductivity (\verb|k|). This dataset only contains a single output-- fuel centerline temperature (\verb|T|). We show a streamlined series of steps used to generate this dataset in Section \ref{sec:results} for comparison with our KAN model.

\subsection{Microreactor Dataset}
The high-temperature gas-cooled reactor (HTGR) dataset models the Holos-Quad HTGR microreactor using Serpent \cite{price_multiobjective_2022}. The reactor features helium gas cooling and graphite moderation. By design, the reactor operates at a maximum of 22 MWth and is controlled via eight cylindrical, materially asymmetrical control drums \cite{price_multiobjective_2022}. As these drums rotate, their boron carbide portions reach a shifting fraction of the core, thus adjusting the power distribution within the reactor as a function of the angles of all eight drums. Accordingly, this dataset contains eight inputs (drum angles denoted \verb|thetaN|, where \verb|N| ranges from 1-8) and four outputs (the flux in each quadrant, denoted \verb|fluxQN| where \verb|N| ranges from 1-4). Each input drum angle ranges from $-180\degree$ to $180\degree$. Though the original dataset only simulated 751 data points, this work leverages reactor symmetry in the four quadrants to expand this number to 3,004 data points. 

\subsection{Power Control Dataset}
This dataset, originally presented by Radaideh et al. \cite{radaideh_neorl_2023}, uses MCNP simulation results based on the Massachusetts Institute of Technology (MIT) Reactor. It seeks to predict the power in each of its 22 fuel elements (thus containing 22 outputs) using the positions of its six control rods (thus containing six inputs). The inputs (control blades labeled \verb|CBN| where N ranges from 1-6) and outputs (power in each fuel element) are denoted based on Figure \ref{fig:MITR_schematic}.

\subsection{Nuclear Safety Dataset}
This dataset, originally presented by Bauer et al. \cite{bauer_results_1993}, seeks to predict time series data for a PWR undergoing a rod ejection accident using PARCS. The data contain four input features: reactivity worth of the ejected rod (\verb|rod_worth|), delayed neutron fraction (\verb|beta|), gap conductance (\verb|h_gap|), and direct heating fraction (\verb|gamma_frac|), and four output features: maximum power (\verb|max_power|), burst width (\verb|burst_width|), maximum fuel centerline temperature (\verb|max_TF|), and average outlet coolant temperature (\verb|avg_Tcool|). This dataset serves as the NEACRP C1 PWR rod ejection accident benchmark, as described in \cite{bauer_results_1993}.

\subsection{Nuclear Cross-Section Dataset}
This dataset, presented by Radaideh et al. \cite{radaideh_shapley_2019} and extended later to more advanced nuclear data modeling with deep neural networks \cite{radaideh2021modeling}, seeks to predict a single output: the multiplication factor ($k_\infty$) of a pressurized water reactor (PWR) fuel assembly. It contains eight input features, all of which are macroscopic cross-sections homogenized over two neutron energy groups-- fast (1) and thermal (2). These features include the following:
\begin{itemize}
    \item $\nu\Sigma_f$ fission for Group 1 (\verb|FissionFast|) and 2 (\verb|FissionThermal|),
    \item $\Sigma_a$ absorption for Group 1 (\verb|CaptureFast|) and 2 (\verb|CaptureThermal|),
    \item $\Sigma_s^{g\rightarrow g}$ in-scattering for Group 1 (\verb|Scatter11|) and 2 (\verb|Scatter22|), 
    \item $\Sigma_s^{1\rightarrow 2}$ down-scattering from Group 1 to Group 2 (\verb|Scatter12|), and 
    \item $\Sigma_s^{2\rightarrow 1}$ up-scattering from Group 2 to Group 1 (\verb|Scatter21|).
\end{itemize} 
Radaideh et al. \cite{radaideh_shapley_2019} generated this simulated dataset using SCALE/TRITON, which builds on fundamental nuclear data libraries that are experimentally determined.

\section{Methodology}
\label{sec:methodology}

Unlike FNNs, KANs lack a standardized methodology for preprocessing, hyperparameter tuning, and post-processing. The KAN demonstrations shown by Liu et al. \cite{liu_kan_2024} provide insight on model architecture, but show primarily ``toy problems''. Liu et al. transfer this to a more generalized \textit{science} approach in their paper \cite{liu_kan_2024-1}, but this work is still largely theory-based. As such, this work scales up and puts previous work into practice by proposing a standardizable methodology for KANs. The KAN models generated in this work rely on \verb|pykan|, an open-source Python package, introduced by Liu et al. \cite{liu_kan_2024}. While the \verb|pykan| package offers a wide variety of à la carte options to polish a KAN, in the name of simplicity and reproducibility, this work focuses on major structural development and leaves accessory features to future work.  

\subsection{Preprocessing}
Preprocessing for any dataset necessarily depends on the structure of that dataset. Thus, we cannot entirely standardize this portion of the modeling process. To prepare the datasets used in this work, we applied some general steps, with slight variations to accommodate each dataset's unique structure. 
\par 
Preprocessing for the KANs in this work begins by loading the datasets and splitting them into input and output columns, followed by a train-test split of 70\%-30\% for training and testing, respectively. We then perform a min-max scaling on both the input ($x$) and output ($y$) data. Following this, the arrays undergo another transformation into \verb|PyTorch| tensors to achieve compatability with the \verb|pykan| package, which is based on \verb|PyTorch|. Finally, in an effort to keep each dataset and its associated information modular and organized, we return a dictionary of the following items for each dataset, packaging it neatly for training: (1) training input data, (2) training output data, (3) testing input data, (4) testing output data, (5) feature (input) labels, (6) output labels, and (7) the $y$ scaler object to un-scale the outputs.

For the Power Control (PC) dataset based on the MIT research reactor, preprocessing also includes splitting the dataset into three sub-regions (A, B, and C) for improved performance. We include further details about this in Section \ref{sec:results}. For the microreactor dataset, we perform a reflection across reactor quadrants as performed and suggested in previous research \cite{myers_pymaise_2025,price_multiobjective_2022}. We also apply a normalization of the fluxes in each quadrant for the microreactor dataset as:
\begin{equation}
    \phi_n = \frac{\phi_n}{\sum_{i=1}^{4} \phi_i}
\end{equation} where $\phi$ is the flux in quadrant $n$. This reflected and normalized dataset, as well as the original, are available in the project repository as described in the \textbf{Data Availability} section. 

\subsection{Hyperparameter Tuning}
\label{sec:hyper_search}

After we complete data preprocessing, we move to hyperparameter tuning. For this process, we found nine key hyperparameters to yield promising results following an in-depth analysis done by the authors on the effect of different parameters. We show these hyperparameters, their descriptions, and the corresponding ranges we tested in Table \ref{tab:hyperparams}.
\par To conduct our hyperparameter tuning, we employed the Python package, \verb|hyperopt|, which requires an objective function, a space, a number of evaluations, and a tuning algorithm \cite{bergstra_making_2012}. We defined the space according to Table \ref{tab:hyperparams}, allowing \verb|hyperopt| to choose each hyperparameter value at random from the list of options given with the exception of $\lambda$ and $\lambda_{entropy}$, which we varied continuously and uniformly over their respective ranges. To define the objective function, we simply built a KAN with the architecture prescribed by the hyperparameters given, and evaluated the model using the average $R^2$ score of its outputs. Since KANs start with their activation functions as splines and then irreversibly convert these splines to symbolic equations, we found average $R^2$ scores for both the spline and symbolic versions of the model, then weighted these results 20-80 for spline and symbolic scores, respectively. Mathematically:
\begin{equation}
    objective = 0.2R_{spline}^2 + 0.8R_{symbolic}^2
\end{equation}
We chose to place significantly higher weight on the symbolic factor ($R^2_{symbolic}$) because we want to optimize for \textit{interpretability} not just accuracy.

We leveraged \verb|hyperopt|'s built-in tuning algorithm, \verb|tpe.suggest|, or plainly, Tree of Parzen Estimators, a type of Bayesian optimization method. We allowed a maximum of 200 evaluations for each dataset, though due to some rare nonfunctional hyperparameter combinations where the model yielded a very poor performance or even failed to train, such combinations are skipped.

\begin{table}[!htbp]
\centering
\caption{Hyperparameter descriptions and ranges for KAN tuning.}
\label{tab:hyperparams}
\begin{tblr}{
  width = \linewidth,
  colspec = {Q[275]Q[383]Q[281]},
  vline{2-3} = {-}{},
  hline{1-2,11} = {-}{},
}
Hyperparameter      & Description                                                                                           & Range                                                                                                                                                                                                                                                                                                                      \\
Depth               & {The number of hidden layers (excluding\\the input and output) in the model.}                         & {[}1, 2]                                                                                                                                                                                                                                                                                                                   \\
Grid                & {Interval range (space) over which the \\model is allowed to build splines.}                          & {[}3, 4, 5, 6, 7, 8, 9, 10]                                                                                                                                                                                                                                                                                                \\
$k$                 & {The polynomial order of the splines used\\as KAN activation functions.}                              & {[}2, 3, 4, 5, 6, 7, 8]                                                                                                                                                                                                                                                                                                    \\
Steps               & {The number of training steps taken by\\pykan.~}                                                      & {{[}25, 50, 75, 100, 125,\\150, 200, 250]}                                                                                                                                                                                                                                                                                 \\
$\lambda$           & {Overall penalty strength used to sparsify\\a KAN.~Ensures activation functions are\\not duplicated.} & 0--0.001, uniformly distributed                                                                                                                                                                                                                                                                                            \\
$\lambda_{entropy}$ & {Entropy penalty strength which is\\ultimately multiplied by \$\textbackslash{}lambda\$.}             & 0--10, uniformly distributed                                                                                                                                                                                                                                                                                               \\
$LR_1$              & {Learning rate used by LFBGS optimizer\\before pruning (on first fit).}                               & {[}0.5, 0.75, 1, 1.25, 1.5, 1.75, 2]                                                                                                                                                                                                                                                                                       \\
$LR_2$              & {Learning rate used by LFBGS optimizer\\after pruning (on second fit).}                               & {[}0.5, 0.75, 1, 1.25, 1.5, 1.75, 2]                                                                                                                                                                                                                                                                                       \\
Reg. Metric         & Regularization type used to sparsify KAN.~                                                            & {{[}``edge\_forward\_spline\_n",\\                                                ``edge\_forward\_sum",\\                                                ``edge\_forward\_spline\_u",\\                                                ``edge\_backward",\\                                                ``node\_backward"]} 
\end{tblr}
\end{table}

\subsection{Symbolic Equation Conversion}

As previously discussed, KANs employ splines as activation functions by default. To enhance interpretability, these splines can be transformed into symbolic equations using available symbolic regression tools. In this work, interpretability and accuracy are both key priorities, and we made specific design choices during the conversion process that may vary in future implementations depending on application needs.

First, we opted to keep the symbolic function library (supported in \verb|pykan|) unrestricted, allowing the conversion process to consider the widest possible range of functions ($\sin$, $\cos$, $\arcsin$, \verb|gaussian|, $x^2$, $x^5$, etc.). We also specify a ``simplicity weight'' parameter during symbolic conversion to control the tradeoff between interpretability and accuracy—where a value of 0 prioritizes accuracy and a value of 1 favors simpler, more interpretable functions. Given the importance of precision in our applications, we selected the lowest simplicity weight, accepting more complex symbolic expressions to preserve model accuracy. Despite their length, these expressions still offer greater transparency compared to the complete black-box nature of traditional machine learning models. To further improve readability, we round all numerical coefficients in the final symbolic equations to four decimal places. Further details about the hyperparameter tuning scheme used in this work can be found in the \verb|hypertuning.py| script in our repository described in the \textbf{Data Availability} section. 

\subsection{KAN Training}
Once we have identified the optimal hyperparameters for a given dataset, we reconstruct the KAN found during hyperparameter tuning. This process consists of initializing a model, fitting it using the limited-memory BFGS (L-BFGS) optimization algorithm \cite{liu1989limited} with the optimal hyperparameter set for the given dataset (using a learning rate of $LR_1$), pruning the model based on default thresholds, and then performing a final refitting using the same hyperparameters but with an updated learning rate of $LR_2$. After the second fitting, we collect metrics from the spline-based model, convert spline activation functions to symbolic functions, then collect a second set of metrics for the symbolic model. At this point, our symbolic equation is ready for explainability analysis. Those seeking to reproduce the results shown in this work can follow instructions in the README available in our repository described in the \textbf{Data Availability} section. 

\subsection{Feedforward Neural Networks}
For comparison against a common black-box algorithm, we also generated feedforward neural networks (FNNs) and associated metrics for each dataset in this study. To obtain the best hyperparameters for each FNN, we used those reported by a previous study \cite{myers_pymaise_2025} that extensively searched for optimal FNN hyperparameters for these datasets using Bayesian optimization. We show these hyperparameters in Table \ref{tab:fnn_hyperparams}. We then applied these hyperparameters to a FNN model architecture and calculated a variety of model metrics for comparison with each dataset's associated KAN. Further details about the FNN model generation used in this work can be found in the \verb|fnn.py| script available in our repository described in the \textbf{Data Availability} section. 

\begin{table}[!htbp]
\centering
\caption{Hyperparameters used for FNN architecture and training as reported by Myers et al. \cite{myers_pymaise_2025}.}
\label{tab:fnn_hyperparams}
\resizebox{\textwidth}{!}{%
\begin{tabular}{lccccccccc}
\toprule
Parameter & CHF & LWR & FP & HEAT & MICROREACTOR & PC* & NS & NXS  \\
\midrule
Hidden Nodes & [231, 138, 267] & [511, 367, 563, 441, 162] & [66, 400] & [251, 184, 47] & [199, 400] & [309] & [326, 127] & [95] \\
Num. of Epochs & 200 & 200 & 200 & 200 & 200 & 200 & 200 & 200 \\
Batch Size & 64 & 8 & 8 & 8 & 8 & 8 & 8 & 8 \\
Learning Rate & 0.000931 & 0.000966 & 0.001000 & 0.000882 & 0.000114 & 0.000832 & 0.000944 & 0.000342  \\
Dropout Layers & True & False & False & False & True & False & False & False \\
Dropout Rate & 0.499590 & 0 & 0 & 0 & 0.322572 & 0 & 0 & 0  \\
\bottomrule
\end{tabular}%
}
\flushleft\footnotesize{*Same hyperparameters were used for the Power Control (PC) problem when training the FNN with the A, B, and C portions of the core.}
\end{table}

\subsection{Explainability Metrics}
Finally, this work would not offer full comprehensibility as described in Section \ref{sec:intro} without an explainability analysis. Since neither a KAN nor FNN is ``explainable'' in its own right, we require supplementary analysis to demonstrate the causational relationships ingrained in each of these models. For this purpose, we employ SHAP, the theory behind which we describe in Section \ref{sec:shap}. To implement SHAP for the KANs, we converted the equation generated for each output into a function (``model'') where SHAP evaluates feature permutations through the symbolic model. As an approximator for the exact SHAP algorithm, Kernel SHAP requires a set of background samples. To determine these background samples, we use a K-means \cite{ahmed2020k} approximation with $k$ clusters/values on the training data according to 0.5\% of the training data multiplied by the number of features in the dataset. If this value for $k$ exceeds 100, we default to $k=100$ to optimize Kernel SHAP runtime. Though Kernel SHAP seems relatively insensitive to sample size and selection method, we chose the K-means approximator in lieu of strict background samples for improved performance, as suggested by the SHAP documentation \cite{lundberg_unified_2017}.  After defining the explainer, we use it to generate SHAP values for all samples contained in the test set and take the mean of the absolute value of those values to determine the average importance of a feature to the overall outcome of the model.

We repeat this process for the FNN, including the definition of the K-means samples, except that instead of a symbolic equation as the ``model," we use the actual FNN model to make predictions on the test set. 

We described previously that classical SHAP is not feasible for magnitude of data points in this model, but we also find it worth noting that Kernel SHAP is not necessarily the best approximation method. Kernel SHAP offers compatibililty with symbolic KAN results because it does not require access to model activation functions (i.e., it offers flexibility to non-neural network models). Future work should attempt to implement more robust SHAP methods to a KAN-FNN comparison analysis. 

To summarize, in this study, we conducted SHAP analyses on both the FNN and symbolic KAN models to consistently evaluate their feature importance rankings, despite the models relying on fundamentally different approaches to generate their predictions.

\subsection{Computational Setup and Resources}

This study employed the following major libraries: pykan-0.2.8, torch-2.6.0, shap-0.46.0, and hyperopt-0.2.7. More details about the package dependencies can be found in our repository described in the \textbf{Data Availability} section.

We conducted the training and hyperparameter tuning on an internal GPU server at the University of Michigan, which is equipped with two AMD EPYC 9654 processors, each providing 96 cores operating at 2.4–3.7 GHz, resulting in a total of 192 cores and 384 threads. Additionally, the server features four NVIDIA RTX 6000 Ada Generation GPUs and 1536 GB of DDR5 RAM.

\section{Results}
\label{sec:results}
Due to the large number of datasets modeled in this work, this section necessarily lacks the entirety of the results we generated. Instead, it focuses on some of the more representative (and interesting) cases, discusses noteworthy findings for all datasets, and leaves the remaining to the Appendices and/or Supplementary Materials. 

\subsection{Hyperparameter Tuning}
\begin{figure}
    \centering
\includegraphics[width=0.7\linewidth]{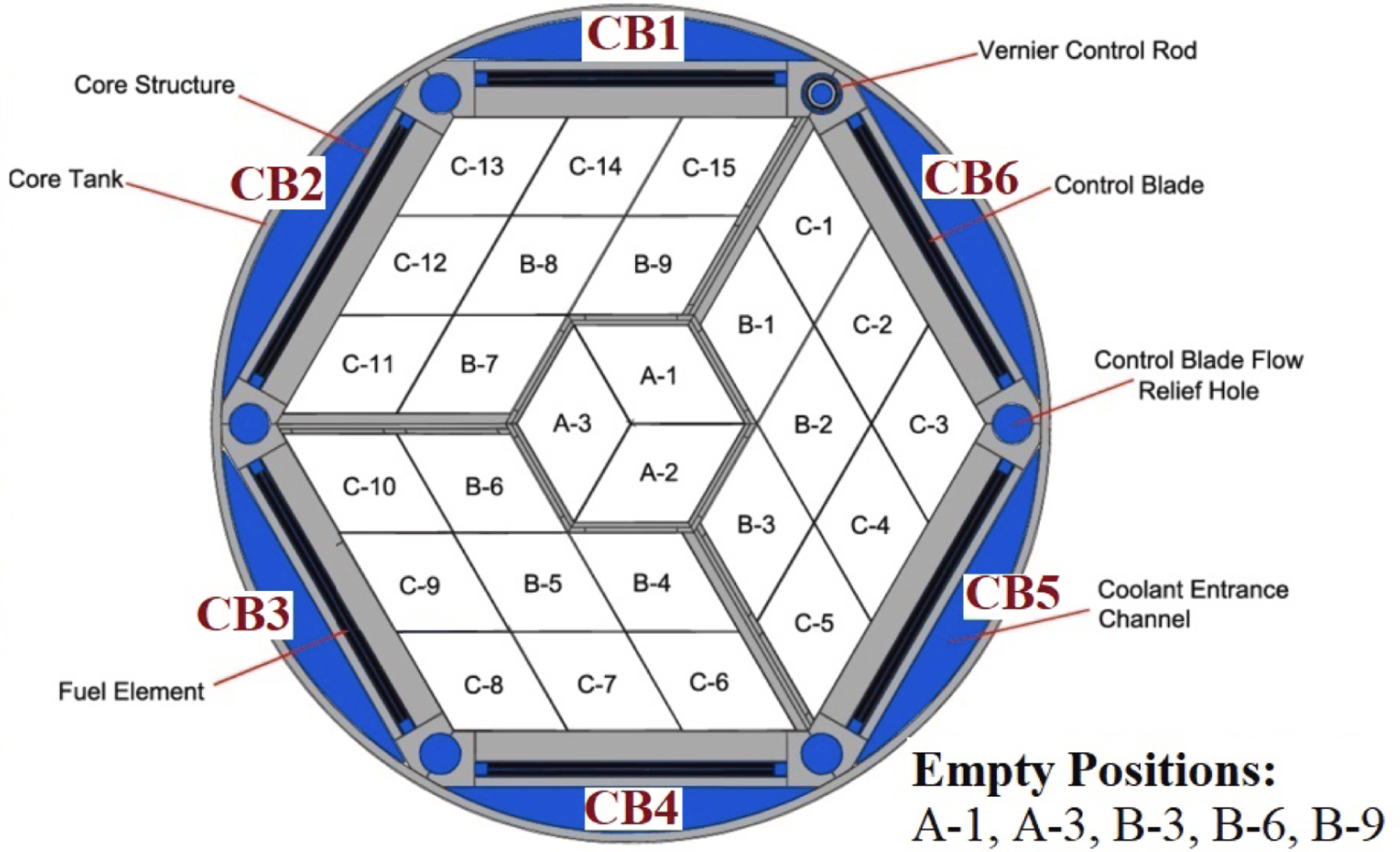}
    \caption{Top-view schematic of the MIT research reactor core used in the Power Control (PC) dataset as shown in \cite{radaideh_neorl_2023}}
    \label{fig:MITR_schematic}
\end{figure}

Table \ref{tab:best_hyperparams} shows the best hyperparameters we found using the search algorithm described in Section \ref{sec:hyper_search} for each dataset, with the exception of the power control (PC) dataset. During training, we found that the power control dataset consistently underperformed the rest of the datasets. This feature is highlighted by the $R^2$ score for \textit{PC} in Table \ref{tab:best_hyperparams} of 0.735. Since this metric diverged from our other datasets' results, we hypothesized that this difference could stem from the larger output space demanded by the power control dataset (22 outputs) which could shed light on KAN performance with high-dimensional output spaces. Accordingly, we split it by physical region of power prediction as shown in Figure \ref{fig:MITR_schematic}. 
This accounts for the datasets abbreviated \textit{PC-A}, \textit{PC-B}, and \textit{PC-C} throughout this work. 

Some abbreviations referenced in hyperparameter tuning results are as follows:
\begin{itemize}
    \item Reg. -- L1 regularization metric used to increase model sparsity,
    \item Spline $R^2$ -- average $R^2$ of all outputs based on model using spline activation functions (before symbolic conversion),
    \item Sym. $R^2$ -- average $R^2$ of all outputs based on model using symbolic activation functions (after symbolic conversion),
    \item EFS -- \verb|edge_forward_sum| (spline and symbolic, norm of the edge using $\frac{\text{output std}}{\text{input std}}$),
    \item EFSU -- \verb|edge_forward_spline_u| (only spline, norm of the edge using output standard deviation),
    \item EFSN -- \verb|edge_forward_spline_n| (only spline, norm of the edge using $\frac{\text{output std}}{\text{input std}}$),
    \item EB -- \verb|edge_backward| (edge attribution score), and
    \item NB -- \verb|node_backward| (node attribution score).
\end{itemize}

Due to the extensive volume of hyperparameter tuning results, we were unable to include the majority of them in this work. However, the tuning process can be replicated using the instructions available in our GitHub repository, as detailed in the Data Availability section. Table \ref{tab:best_hyperparams} gives some insight on the overall KAN performances with the best hyperparameter configuration found from the search based on the final symbolic and spline-based $R^2$ results for each dataset. As previously mentioned, the full power control dataset (PC) core performs the worst, followed by the light water reactor (LWR) dataset, with a symbolic $R^2$ of 0.90596. The rest of the datasets approximately range between 0.97 to 0.99, with most residing on the upper end of that spectrum. We will discuss potential causes for weaker performance in Section \ref{sec:discussion}. Overall, the symbolic $R^2$ closely matches the spline $R^2$ across all problems, regardless of the $R^2$ value. This indicates that the interpretable version of KAN maintains its predictive accuracy. 

\begin{table}[!htbp]
\centering
\caption{Best KAN hyperparameter tuning results for all datasets.}
\label{tab:best_hyperparams}
\resizebox{\textwidth}{!}{%
\begin{tabular}{lccccccccccc}
\toprule
Dataset & Depth & Grid & $k$ & $\lambda$ & $\lambda_{entropy}$ & $LR_1$ & $LR_2$ & Reg. Metric & Steps & Spline $R^2$ & Symbolic $R^2$ \\
\midrule
FP      & 1 & 8 & 7 & 2.043e-05 & 5.03464  & 1.5  & 1.75  & EFS  & 75  & 0.99144 & 0.99098 \\
LWR     & 1 & 7 & 2 & 8.912e-04 & 7.48809  & 1.75 & 1.25  & EFS  & 125 & 0.96004 & 0.90596 \\
HEAT    & 1 & 7 & 3 & 1.899e-04 & 8.20921  & 1.5  & 2     & EFSU & 150 & 0.99308 & 0.99823 \\
MICROREACTOR    & 1 & 8 & 3 & 1.217e-05 & 7.66581  & 0.75 & 1.25  & EFS  & 25  & 0.99183 & 0.98926 \\
PC-A  & 1 & 4 & 7 & 4.284e-05 & 0.02540  & 1.5  & 1.25  & EFS  & 25  & 0.99859 & 0.99843 \\
PC-B  & 1 & 3 & 2 & 1.112e-06 & 0.96268  & 0.75 & 1     & EFS  & 250 & 0.99766 & 0.99745 \\
PC-C  & 1 & 3 & 3 & 9.739e-06 & 7.04743  & 1    & 1.5   & EFSN & 150 & 0.97215 & 0.97204 \\
PC - Full Core    & 1 & 7 & 6 & 3.370e-04 & 5.77322  & 1.75 & 0.5   & EFSU & 150 & 0.74729 & 0.73557 \\
CHF     & 1 & 9 & 2 & 5.123e-06 & 7.71662  & 2    & 1.75  & EFSU & 100 & 0.99675 & 0.99118 \\
NS     & 1 & 6 & 8 & 3.795e-05 & 0.00650  & 2    & 0.5   & EFS  & 25  & 0.99681 & 0.99555 \\
NXS      & 2 & 9 & 4 & 3.903e-04 & 0.42861  & 1.25 & 1.25  & EFSU & 100 & 0.99988 & 0.99993 \\
\bottomrule
\end{tabular}%
}
\end{table}

\subsection{Metrics}
The three models our analysis will focus on from here includes the light water water reactor dataset (LWR), the heat conduction dataset (HEAT), and the critical heat flux dataset (CHF). Tables \ref{tab:chf_metrics}, \ref{tab:heat_metrics}, and \ref{tab:bwr_metrics} present a variety of model metrics for both the best KANs and FNNs for the CHF, HEAT, and LWR datasets, respectively. \textit{For clarity, at this point in the analysis, we will refer to the symbolic KAN as plainly, the KAN, which is the interpretable KAN version of interest}. 

After hyperparameter tuning, we convert all KAN model activation functions from splines to symbolic equations to analyze the most interpretable form of the model. The metrics included in Tables \ref{tab:chf_metrics}, \ref{tab:heat_metrics}, and \ref{tab:bwr_metrics} include: the mean absolute error (MAE), the mean absolute percentage error (MAPE), the mean squared error (MSE), the root mean squared error (RMSE), the root mean squared percentage error (RMSPE), and the symbolic $R^2$ (denoted as strictly $R^2$). \textit{All these metrics are reported for the withheld test dataset}. 

\begin{table}[htbp]
\centering
\caption{Performance metrics based on the test set of the CHF dataset as modeled by both a \textbf{symbolic} KAN and FNN.}
\label{tab:chf_metrics}
\begin{tabular}{llrrrrrr}
\toprule
Output & Model & MAE & MAPE (\%) & MSE & RMSE & RMSPE (\%) & $R^2$ \\
\midrule
CHF  & KAN & 87.85040  & 7.09290  & 17150.37790  & 130.95950  & 12.87050  & 0.99010 \\
CHF  & FNN & 22.19988  & 1.49471  & 1196.16250  & 34.58558   & 2.11510   & 0.99931 \\
\bottomrule
\end{tabular}%
\end{table}

Table \ref{tab:chf_metrics} shows all model metrics for the best KAN and FNN for the CHF dataset. As shown in Table \ref{tab:chf_metrics}, the FNN outperforms the KAN just slightly by 0.009 in $R^2$, with both models performing extremely well in terms of other metrics. As a reference to a common correlation used to predict CHF, Eq.\eqref{eq:W3CHF} shows the W-3 (Westinghouse 3) correlation used to predict CHF for the departure from nuclear boiling (DNB) condition \cite{buongiorno_shape_2010}. For comparison, Eq.\eqref{eq:CHF} shows the symbolic equation generated from the KAN for CHF. 

\begin{multline}
\label{eq:W3CHF}
CHF_{W-3} =\left\{(2.022-0.06238 \mathit{P})+(0.1722-0.01427 \mathit{P}) \exp \left[(18.177-0.5987 \mathit{P}) \mathit{X}_{\mathit{e}}\right]\right\} \\
{\left[\left(0.1484-1.596 \mathit{X}_{\mathit{e}}+0.1729 \mathit{X}_{\mathit{e}} \mid \mathit{X}_{\mathit{e}}\right) \cdot 2.326 \mathit{G}+3271\right] \cdot\left[1.157-0.869 \mathit{X}_{\mathit{e}}\right]} \\
{\left[0.2664+0.8357 \exp \left(-124.1 \mathit{D}_{\mathit{h}}\right)\right] \cdot\left[0.8258+0.0003413\left(\mathit{~h}_{\mathit{f}}-\mathit{h}_{\text {in }}\right)\right],}
\end{multline}
where $h$, the only variable not considered in our CHF dataset, represents enthalpy and $D_h$ specifies a heated diameter. Note that in KAN, we used the inlet temperature $T_{in}$ instead of the enthalpy as reported in the CHF dataset, where both are redundant (e.g., inlet enthalpy can be inferred by the temperature). The following is the KAN's CHF symbolic equation: 

\small{
\begin{multline}
\label{eq:CHF}
    CHF_{KAN} = 0.0358 \left(- \tan{\left(1.4154 \mathit{T_{in}} + 2.5774 \right)} + 0.9336 \tanh{\left(4.4497 \mathit{Xe} - 1.6801 \right)}+ 0.8381 \operatorname{asin}{\left(1.764 \mathit{P} - 0.999 \right)} - 0.4316\right)^{5} \\
    + 0.0233 \left(0.4465 \log{\left(4.2995 - 3.7156 \mathit{T_{in}} \right)} - 0.4027 \log{\left(1.6 \mathit{L} + 0.004 \right)} + 0.39 \log{\left(7.0 \mathit{G} + 0.02 \right)} + \alpha_1 + \alpha_2 - \frac{1}{\left(- 0.815 \mathit{D} - 1\right)^{2}}\right)^{4} \\
    - 0.4537 \left|{0.0235 \tanh{\left(9.6 \mathit{G} - 2.052 \right)} + 0.018 \operatorname{atan}{\left(9.7986 \mathit{Xe} - 6.34 \right)} + 0.0061}\right| + 0.0145,
\end{multline}}

where:
\begin{align*}
\alpha_1 &= 0.1789 \operatorname{atan}{\left(3.7926 \mathit{Xe} - 1.9798 \right),} \\
\alpha_2 &= 0.0969 \operatorname{atanh}{\left(1.956 \mathit{P} - 0.9857 \right)} + 0.8565. \\
\end{align*}

Note that the coefficients $\alpha_1$ and $\alpha_2$ were defined to allow the equation to fit within the space. Clearly, the KAN-generated equation, Eq.\eqref{eq:CHF}, is longer and more complicated than the W-3 correlation in Eq.(\ref{eq:W3CHF}), but the KAN-generated equation also relies on more complex functions as artifacts of the B-splines used to fit the KAN. This makes the KAN-generated equation more difficult to understand at first glance, i.e., rendering it less interpretable than the correlation. Worth note, however, is the fact that the W-3 correlation is only reliable under certain parameter ranges \cite{buongiorno_shape_2010}. While we could also argue the KAN-equation may only be valid under the range of the training set, we could expand the training data to quickly adapt a KAN to fit a given application, which is the case for this CHF dataset that is more diverse and comprehensive than the W-3 correlation. Thus, the KAN offers more flexibility than existing correlations. Though a thorough comparison between these two equations resides outside the scope of this paper, relating a KAN-generated equation to known correlations for CHF or other applications could present an interesting future work.

\begin{table}[!htbp]
\centering
\caption{Performance metrics based on the test set of the heat conduction (HEAT) dataset as modeled by both a \textbf{symbolic} KAN and FNN.}
\label{tab:heat_metrics}
\begin{tabular}{llrrrrrr}
\toprule
Output & Model & MAE & MAPE (\%) & MSE & RMSE & RMSPE (\%) & $R^2$ \\
\midrule
T  & KAN & 7.18120  & 0.56590  & 77.89160  & 8.82560  & 0.70590  & 0.99630 \\
T  & FNN & 9.07959  & 0.70132  & 1004.82886 & 31.69904 & 2.44894  & 0.95280 \\
\bottomrule
\end{tabular}%
\end{table}

As shown by Table \ref{tab:heat_metrics}, the KAN model of the heat conduction dataset outperforms the FNN for all performance metrics, specifically scoring about 0.044 higher on $R^2$. This dataset boasted the advantage of only predicting a single output compared with many other datasets in this section, but was of particular interest to this analysis because the dataset itself is based on a numerical model for solving the heat condition based on a quasi-two-dimensional analysis (a.k.a 1.5D  conduction) \cite{williamson_multidimensional_2012}. While the steps to generate this dataset do not boil down to a single algebraic equation, the process illuminates some known dependencies and relationships between the input features. We show the steps used to generate the heat conduction data as follows: Given a temperature-dependent thermal conductivity modeled as a polynomial
\begin{equation}
k(T) = A T^3 + B T^2 + C T + D
\end{equation}
The integral of \( k(T) \) with respect to temperature is:
\begin{equation}
\int k(T)\, dT = \frac{A}{4} T^4 + \frac{B}{3} T^3 + \frac{C}{2} T^2 + D T
\end{equation}
At a given axial location \( z \), the reference temperature at the outer radius \( r = R \) is computed as:
\begin{equation}
T_{\text{ref}}(z) = \frac{q'}{\dot{m} C_p} z + T_{\text{in}}
\label{eq:Tref}
\end{equation}
A constant is then defined at each radial position \( r \) and axial location \( z \) as:
\begin{equation}
\text{const}(r, z) = \frac{q'}{4\pi} \left(1 - \left( \frac{r}{R} \right)^2 \right) + \int k(T_{\text{ref}}(z))\, dT
\end{equation}
To determine the temperature at radius \( r \), the following nonlinear equation is solved:
\begin{equation}
\frac{A}{4} T^4 + \frac{B}{3} T^3 + \frac{C}{2} T^2 + D T - \text{const}(r, z) = 0
\end{equation}
This equation is solved numerically, with the temperature at the next outer radial node used as the initial guess. Then we aim to solve for $T(r,z)$ for every value of $r$ and $z$  using root finding algorithms (e.g., Levenberg–Marquardt method) such that the temperature $T$ satisfies the following:
\begin{equation}
\frac{A}{4} T^4 + \frac{B}{3} T^3 + \frac{C}{2} T^2 + D T - \frac{q'}{4\pi} \left(1 - \left( \frac{r}{R} \right)^2 \right) + \int k(T_{\text{ref}}(z))\, dT  = 0
\label{eq:heat_final_numerical}
\end{equation}

Based on the equations shown for this process, one would expect $q'$ and $k$ at least, to strongly influence the final output, as these variables serve as coefficients in Eq.\eqref{eq:heat_final_numerical}. Similarly, we might also expect some smaller contributions from $\dot m$, $C_p$, and $T_{in}$, as shown in Eq.\eqref{eq:Tref}. In contrast, we show the equation generated by KAN in Eq.\eqref{eq:HEAT_KAN} as

\begin{multline}
\label{eq:HEAT_KAN}
    T = - 4.0045 \left( 0.0051 \left( - 0.1461 \mathit{L} - 1 \right)^{5} 
    - 0.0122 \tan{\left( 2.3733 \mathit{q'} + 8.217 \right)} \right. \\
    + 0.0021 \tan{\left( 2.291 \dot m - 1.2526 \right)} 
    - 0.0231 \operatorname{atan}{\left( 1.521 \mathit{k} - 0.6516 \right)} \\
    \left. - 1 \right)^{3} 
    + 0.1707 \operatorname{sign}{\left( - 1.3229 \operatorname{sign}{\left( 4.904 - 9.0 \mathit{q'} \right)} \right.} \\
    \left. - 0.0077 + 0.1071 e^{- 100.0 \left( 0.38 - \mathit{L} \right)^{2}} \right) - 3.6878
\end{multline}

While a detailed comparison between these two solutions is hardly obvious to the naked eye, one feature-- the complexity-- is. Contrary to simpler correlations used to predict CHF, Eq.\eqref{eq:HEAT_KAN}, coupled with impressive performance metrics highlights the interpretability yielded by this ML method. As verification, we can also see several of the variables we expected from the numerical calculation featured in Eq.\eqref{eq:HEAT_KAN}. 

As shown by Table \ref{tab:bwr_metrics}, based on the $R^2$, the FNN outperforms the KAN for the light water reactor dataset for all outputs except for Max-Fxy, which performs worst among all outputs between both models. KAN scores 0.77960 for this output, while FNN only scores 0.47181, the largest difference between the two models. The next highest performance gap for this dataset is only about 0.06. The average $R^2$ for FNN for this dataset is 0.87798, coming in slightly under the $R^2$ symbolic average for KAN, which was 0.90596. Overall, this was the worst performing dataset for both models without alterations (the power control ``full core'' dataset underperformed significantly for KAN before being subdivided by core region). It is worth noting that other metrics provided for both KAN and FNN results for Max-Fxy predictions on this dataset indicate that the difference in poor performance between the KAN and the FNN is magnified by $R^2$, while MAE and MAPE, in particular, show more comparable metrics, implying both KAN and FNN are comparable.

\begin{table}[!htbp]
\centering
\caption{Performance metrics for the light water reactor (LWR) dataset as modeled by both a \textbf{symbolic} KAN and FNN based on the test set.}
\label{tab:bwr_metrics}
\begin{tabular}{llrrrrrr}
\toprule
Output & Model & MAE & MAPE (\%) & MSE & RMSE & RMSPE (\%) & $R^2$ \\
\midrule
F-delta-H & KAN & 0.03290 & 2.11260 & 0.00210 & 0.04600 & 2.90030 & 0.95020 \\
F-delta-H & FNN & 0.01753 & 1.10367 & 0.00077 & 0.02776 & 1.65984 & 0.98188 \\
\midrule
K-eff     & KAN & 0.00920 & 1.07560 & 0.00020 & 0.01380 & 1.76650 & 0.97580 \\
K-eff     & FNN & 0.00384 & 0.44452 & 0.00005 & 0.00733 & 0.90591 & 0.99317 \\
\midrule
Max-Fxy   & KAN & 0.01460 & 0.80130 & 0.00050 & 0.02240 & 1.26480 & 0.77960 \\
Max-Fxy   & FNN & 0.01342 & 0.74877 & 0.00120 & 0.03471 & 2.05290 & 0.47181 \\
\midrule
Max3Pin   & KAN & 0.31330 & 7.79620 & 0.22630 & 0.47580 & 11.44530 & 0.91190 \\
Max3Pin   & FNN & 0.11275 & 2.58399 & 0.07184 & 0.26803 & 6.68874 & 0.97204 \\
\midrule
Max4Pin   & KAN & 0.33070 & 7.92020 & 0.26010 & 0.51000 & 11.60500 & 0.91230 \\
Max4Pin   & FNN & 0.11693 & 2.59271 & 0.08609 & 0.29340 & 6.98493 & 0.97099 \\
\bottomrule
\end{tabular}%
\end{table}

Eq.\eqref{eq:f-delta-h_BWR} represents the symbolic expression generated by the KAN for the light water reactor (LWR) dataset, specifically for the output F-delta-H, which is chosen for demonstration. At first glance, this expression reveals that KAN can also produce highly complex formulas, making interpretation challenging, as it requires defining seven $C$ terms just to fit within the page margins. However, despite its complexity, the KAN expression remains more manageable compared to the stacked arrays of weights and biases used in FNNs. 

\begin{multline}
    \label{eq:f-delta-h_BWR}
    \text{F-delta-H} = - 0.095 \operatorname{atan}{\left(34.6246 \left(0.5079 - \mathit{FlowRate}\right)^{5} + 6.1256 \left(0.52 - \mathit{subcool}\right)^{3} - C_1 + 1.27 + 0.1551 e^{C_2} - 3.2746 C_7 \right)} \\ + 0.1085 + 0.1906 e^{C_3} + 0.4772 e^{ - 12.1176 (C_4 + C_6)^{2}}
\end{multline}

where:

\begin{align*}
C_1 &= 1.3814 \operatorname{atan}{\left(4.652 \mathit{vanA} - 3.4542 \right)} - 1.8174 \operatorname{atan}{\left(2.9934 \mathit{vanB} - 2.1584 \right)} \\
C_2 &=  - 100.0 \left(0.64 - \mathit{PowerDensity}\right)^{2} \\ 
C_3 &= - 2.7504 \left(- 0.2044 \tanh{\left(5.69 \mathit{vanB} - 3.6418 \right)} + \tanh{\left(1.38 \mathit{CRD} - 1.3628 \right)} - 0.234 \operatorname{asin}{\left(1.3781 \mathit{DOM} - 0.7163 \right)} - C_5\right)^{2} \\
C_4 &= - 0.2481 \sin{\left(1.1506 \mathit{DOM} - 6.7567 \right)} - 0.6109 \operatorname{acos}{\left(- 0.4426 \mathit{PSZ} - 0.156 \right)} \\
C_5 &= 0.1458 \operatorname{atan}{\left(9.906 \mathit{vanA} - 7.216 \right)} + 0.5282 + 0.4733 e^{- 0.6815 \left(- \mathit{PSZ} - 0.4916\right)^{2}} \\
C_6 &= 0.0393 \operatorname{atan}{\left(5.4866 \mathit{vanA} - 3.9469 \right)} + 0.0573 \operatorname{atan}{\left(2.4404 \mathit{vanB} - 1.346 \right)} + \operatorname{atan}{\left(1.6189 \mathit{CRD} + 0.1474 \right)} + 0.7047 \\
C_7 &= e^{- 23.04 \left(0.3192 - \mathit{CRD}\right)^{2}}
\end{align*}

Note that the variable $C_5$ is used in $C_4$ calculations and does not appear in the main Eq.\eqref{eq:f-delta-h_BWR}. The key question now is whether F-delta-H can be estimated more simply using mathematical formulas. F-delta-H accounts for axial power peaking effects, ensuring that hot nuclear channels remain within safe limits. It is defined as the ratio of the maximum enthalpy rise in all channels to the average, expressed as:
\begin{equation}
\label{eq:fdeltah_physics}
F_{\Delta H} = \frac{\max \left( \int_0^L q' (z) \, dz \right)}{\frac{1}{N} \sum\limits_{i=1}^{N} \int_0^L q'_i (z) \, dz}
\end{equation}
where $q'(z)$ is the linear heat generation rate at axial position $z$ with unit (W/cm), $L$ is the active fuel length (cm), and $N$ is the total number of fuel channels in the core (in our case, for a mini-core, there are 4 channels). 

At first glance, this may seem like a simple ratio compared to the complex KAN formula in Eq. \eqref{eq:f-delta-h_BWR}. However, a closer examination reveals that estimating the power distribution through the function $q'(z)$ across all channels in the reactor core is a highly intricate task. It requires solving the neutron transport equation, which implicitly depends on various operating parameters and boundary conditions such as subcooling, CRD position, flow rate, power density, and VFNGAP. The axial height of different fuel zones—PSZ, DOM, vanA, and vanB—is implicitly embedded in the integration limit $L$, representing the total fuel height. Determining the power distribution is the heart of nuclear reactor analysis and typically relies on legacy computational codes that, despite being based on numerical methods for solving partial differential equation systems, are often considered "black-box" solutions due to their complexity. In this study, we used SIMULATE3, a widely popular nuclear simulation code, to estimate $q'(z)$ and then F-delta-H based on the perturbation of the nine inputs provided to KAN in Eq.\eqref{eq:f-delta-h_BWR}. Thus, the key takeaway is that the KAN-derived equation, despite its complexity, remains significantly more interpretable and manageable compared to the ``black-box'' nuclear code estimation of F-delta-H. 

At this point, the attentive reader might notice that the symbolic $R^2$ metrics presented in Tables \ref{tab:chf_metrics}, \ref{tab:heat_metrics}, and \ref{tab:bwr_metrics} vary slightly from those presented in Table \ref{tab:best_hyperparams}. These small differences are caused by the stochastic nature of fitting the KANs and converting their splines to symbolic equations. Even with the same hyperparameters and seed (defaulted to 42 for all models in this work), we still observe some minor variations between the models generated during hyperparameter tuning and those recreated during the full analysis.

\subsection{Explainability}
\begin{figure}[!h]
\centering
\label{fig:bwr_explain}
\begin{subfigure}[b]{1\textwidth}
   \includegraphics[width=1\linewidth]{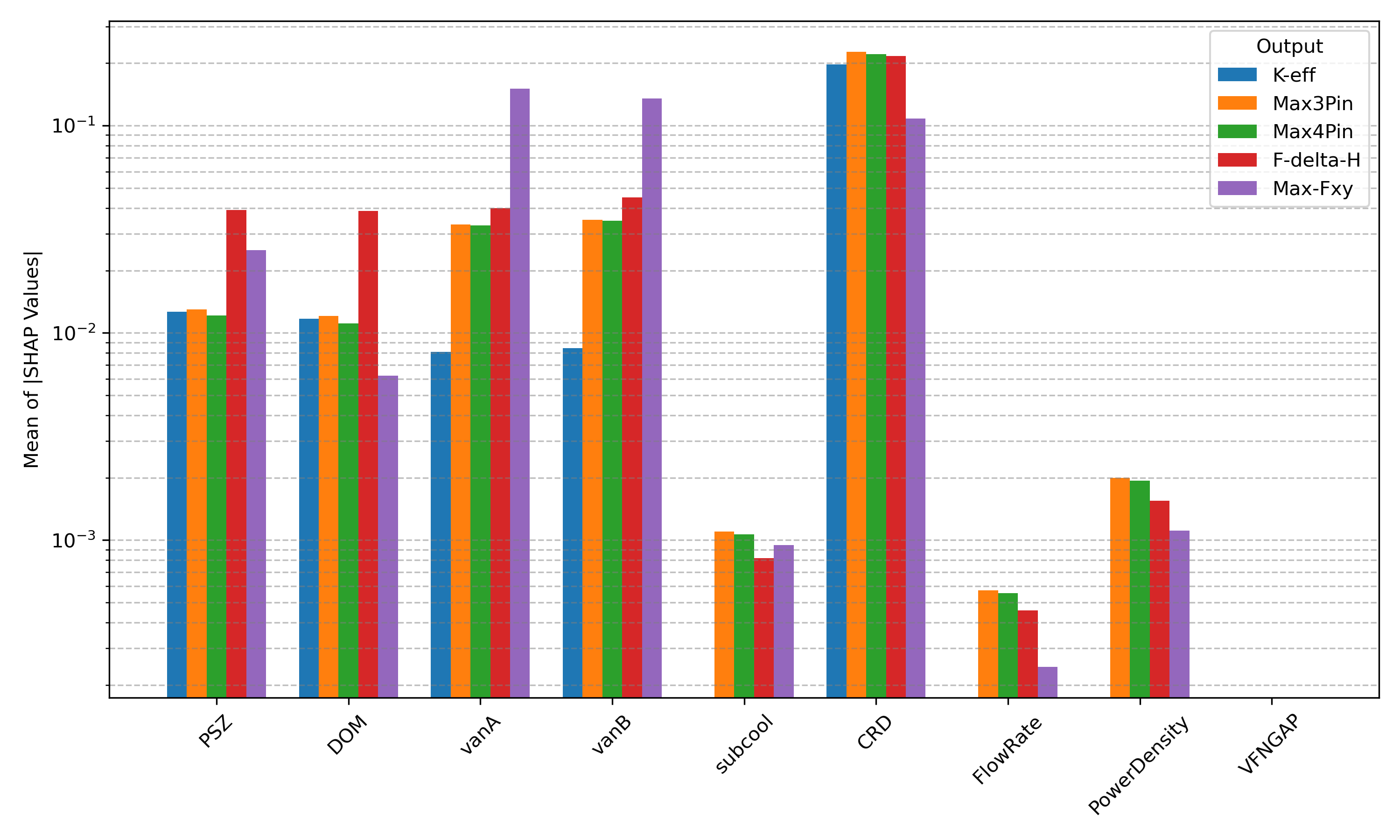}
   \caption{Symbolic KAN}
   \label{fig:bwr_kan_xai} 
\end{subfigure}
\begin{subfigure}[b]{1\textwidth}
   \includegraphics[width=1\linewidth]{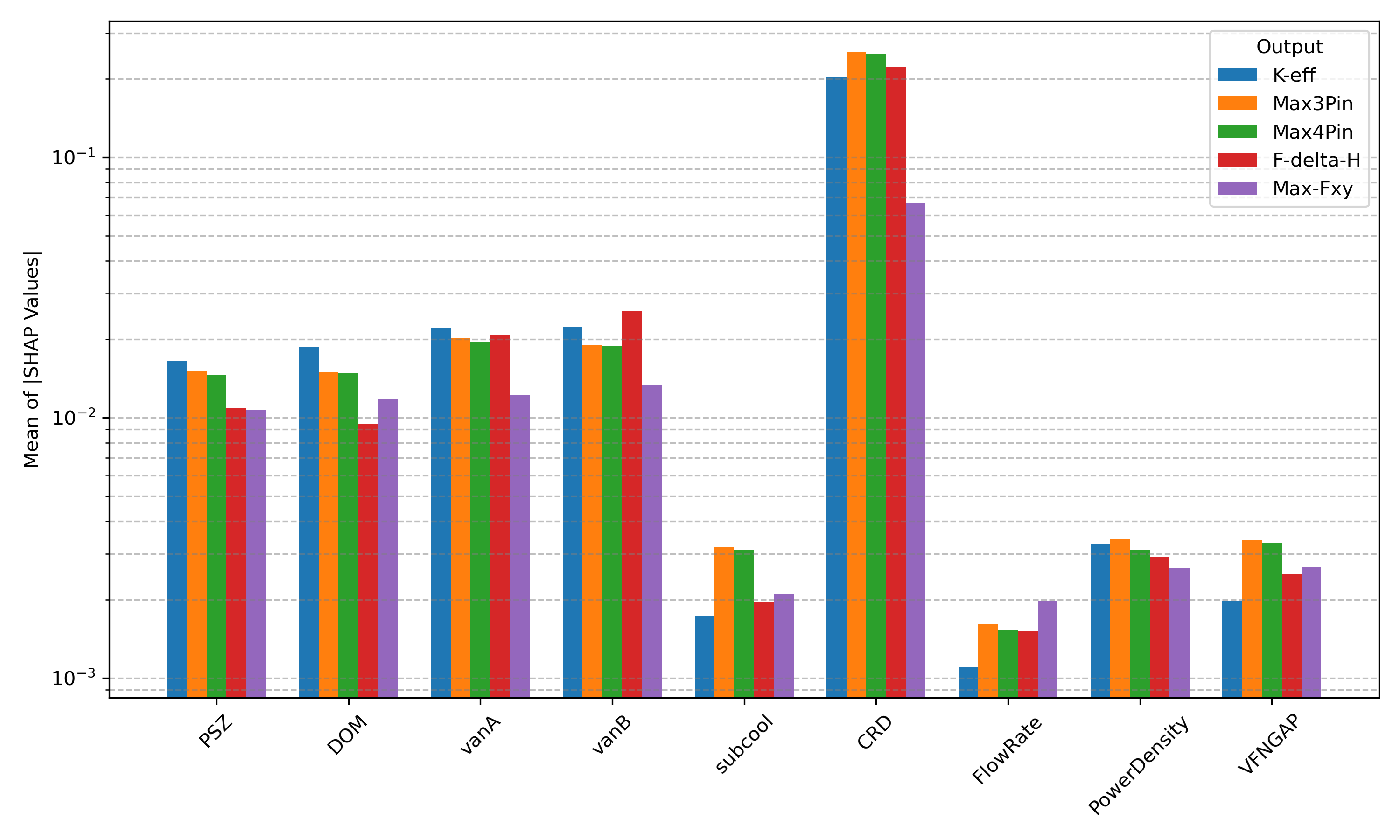}
   \caption{FNN}
   \label{fig:bwr_fnn_xai}
\end{subfigure}

\caption[LWR Explainability Results]{Feature importance ranking via SHAP for all outputs for both models of the light water reactor (LWR) dataset.}
\end{figure}

We show the results of the Kernel SHAP explainability analysis we conducted on the light water reactor (LWR) dataset for both the KAN in Figure \ref{fig:bwr_kan_xai} and the FNN in Figure \ref{fig:bwr_fnn_xai}. Overall, both models largely agreed with the general importance rankings for each of the input features, but they varied on an individual output basis. For example, the KAN ranked the importance of PSZ highest for the F-delta-H output between other outputs, but the FNN ranked that feature second lowest between other outputs and the PSZ's feature attribution to them. Yet, both the KAN and FNN found CRD to be the most important feature for predicting all outputs. This is expected since CRD represents the control rod insertion which affects the power distribution significantly in the reactor. These individual variations give us insight as to how much the FNN and the KAN are each using the given features in their prediction method. We can glean even more intuition for the KAN by studying its output equation, but unfortunately, these explainability metrics approach the upper limit of our insight into the FNN. 

\begin{figure}[!h]
    \centering
    \includegraphics[width=0.6\linewidth]{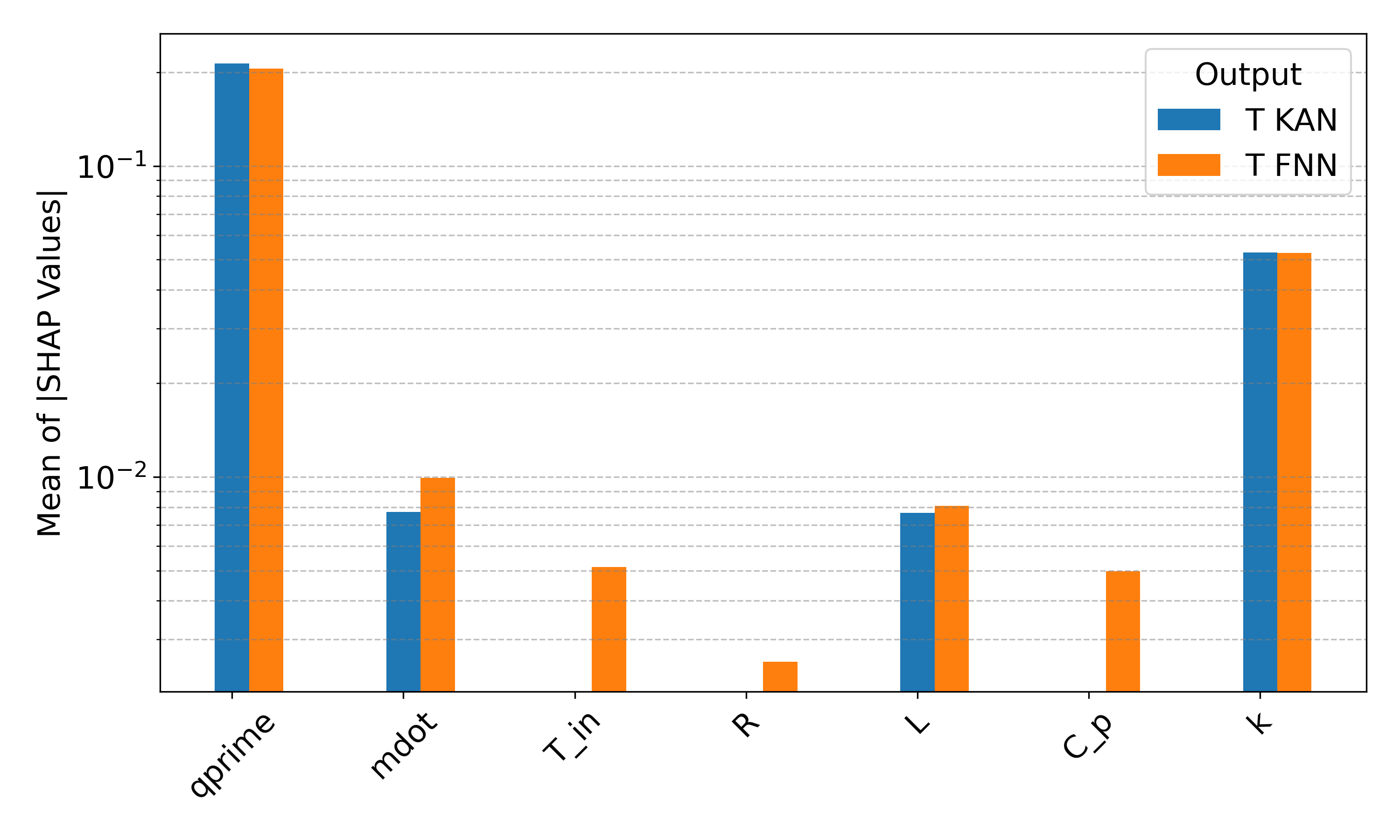}
    \caption{Feature importance ranking via SHAP for all outputs for both symbolic KAN and FNN models of the heat conduction (HEAT) dataset.}
    \label{fig:heat_explain}
\end{figure}

We show the feature importance ranking for the given heat conduction (HEAT) dataset input features for both the KAN and the FNN in Figure \ref{fig:heat_explain}. Figure \ref{fig:heat_explain} is particularly interesting because it shows that the KAN only used four of seven input features to make its predictions, even though we know that those input features were used to generate the dataset. Further, even though the FNN used all input features to make its predictions, it still performs worse than the KAN. The KAN and the FNN do both agree on the two most important features for predicting temperature: qprime ($q'$) and thermal conductivity ($k$) as indicated before in Eq.\eqref{eq:heat_final_numerical}-\eqref{eq:HEAT_KAN}.

\begin{figure}[!h]
    \centering
    \includegraphics[width=0.6\linewidth]{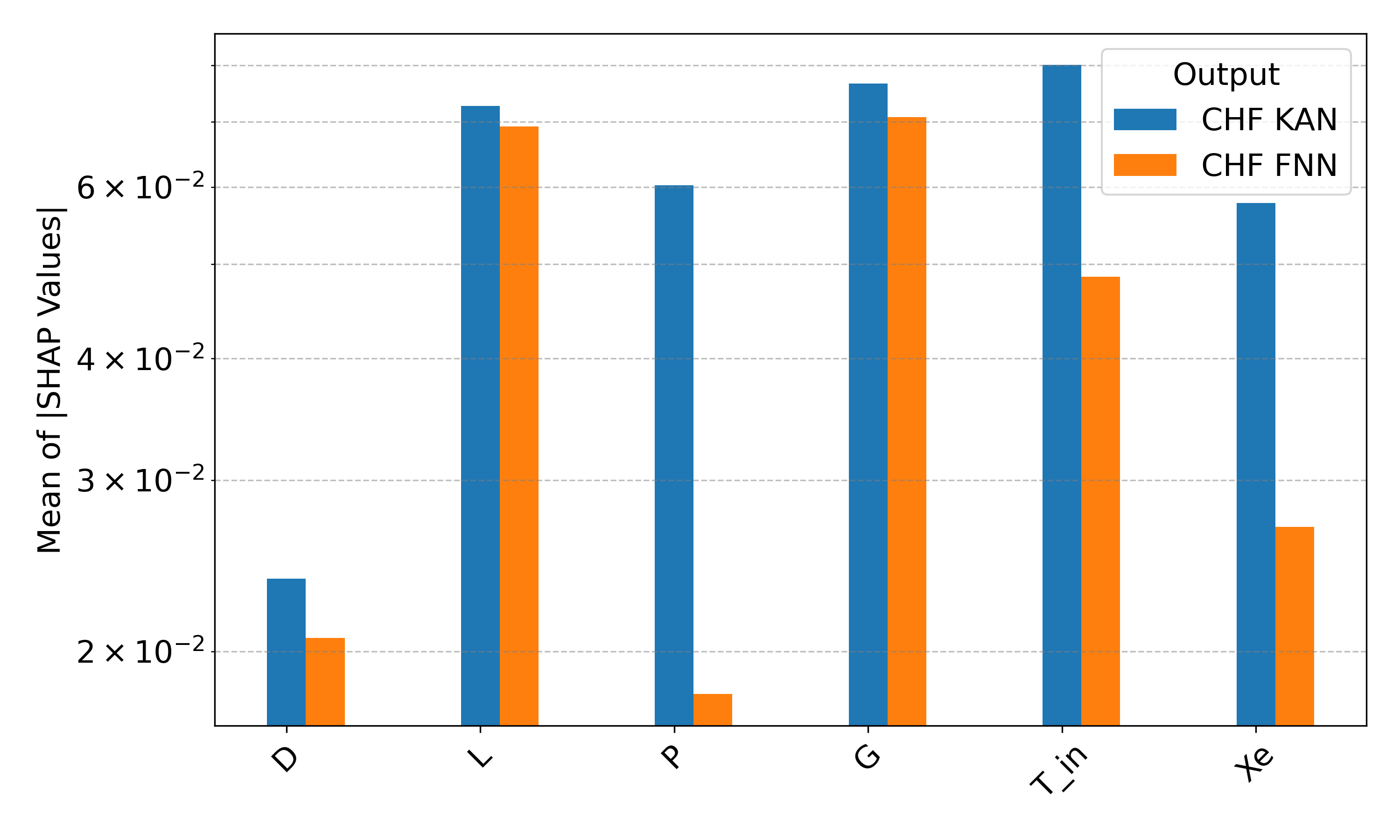}
    \caption{Feature importance ranking via SHAP for all outputs for both symbolic KAN and FNN models of the critical heat flux (CHF) dataset.}
    \label{fig:chf_explain}
\end{figure}

Finally, we show the Kernel SHAP explainability metrics for the CHF dataset for the KAN and FNN in Figure \ref{fig:chf_explain}. In a result somewhat flipped from the heat conduction dataset, the KAN actually weighs more inputs as ``important'' compared to the FNN. We especially see this impact on pressure ($P$) and outlet quality ($Xe$), where both have a mean absolute SHAP value of nearly 0.06 for the KAN, while $Xe$ and $P$ have a mean absolute SHAP values of about 0.0275 and 0.01 for the FNN, respectively. Since we know that pressure and outlet quality are thermodynamically linked quantities, it appears that the KAN gives them both importance, while the FNN uses one over the other, with both models yielding reliable prediction results. We also observe that the most important feature for the KAN CHF model is inlet temperature ($T\_in$) whereas it is mass flux ($G$) for the FNN model. Additionally, the FNN places significantly less importance on T\_in and slightly more emphasis on the channel diameter ($D$). 

\subsection{Other Models}
Due to space limitations, we will summarize the key takeaways of the remaining datasets tested in this work here. We show all dataset metrics in  \ref{sec:metrics}, all explainability plots in \ref{app:shap-plots}, while \ref{app:equations} provides more instructions on how to access all symbolic equations in the \textbf{Supplementary Materials}, which are tedious to type and fit in this paper. In general, the KAN and the FNN were comparable in performance for all datasets. KAN marginally outperformed FNN for the fuel performance and microreactor datasets, while FNN outperformed KAN significantly for the power control full core dataset. Once we spatially divided the power control dataset, KAN and FNN performance became much more comparable across all metrics. We found KAN and FNN performance comparable across the nuclear cross section and the nuclear safety datasets.

Further, we generally saw agreement between the SHAP feature importance rankings across all datasets, with some small exceptions. In general, we found KANs more likely to have mean absolute SHAP scores of 0, meaning KANs are more likely to ``eliminate'' a variable (and thereby reduce the complexity of their problem space) than a KAN. We observe this phenomenon particularly well for the nuclear cross section dataset shown in Figures \ref{fig:xs_kan} and \ref{fig:xs_fnn}, respectively. Both of these models produced $R^2$ scores of greater than 0.999, but the KAN predicts with this accuracy using three less input features than the FNN. These three features (scatter\_11, scatter\_21, scatter\_22) are known from the physics of this problem to have a minimal impact on the multiplication factor (i.e., output). This is due to the low likelihood of upscattering or in-group scattering, which KAN appears to fully capture and represent.  

Finally, the modeling results provided in this work are reproducible by setting the KAN hyperparameter \verb|seed=42|. While the seed can impact model results, we tested a variety of seed values during our analysis and found no major changes to our results, even using the hyperparameters optimized for \verb|seed=42|. Those looking to use a different seed for the datasets used in this work, though, \textit{should} repeat hyperparameter tuning to ensure optimal model performance.

\section{Discussion}
\label{sec:discussion}
Throughout this work, we place emphasis on the following three features of our models: accuracy, interpretability, and explainability. In this section, we will discuss our results through these three lenses and consider the successes, as well as shortcomings, of KANs for the diverse datasets used in this work. We will then discuss some of the practical implications of this work on using KAN for regression problems. We will also dive into the three datasets highlighted in Section \ref{sec:results} and some insights we gleaned from these models. 

\subsection{Accuracy}
In Section \ref{sec:results}, we describe the model metrics and explainability analysis results for the light water reactor (LWR) dataset. We chose this dataset to highlight due to its poor performance between both the KAN and the FNN, particularly when predicting the output Max-Fxy. From $R^2$ alone, it appears that the KAN outperforms FNN substantially in this area, but as previously described, the other metrics provide more comparable statistics. Of all the datasets evaluated in this work, only the light water reactor and power control cases showed large differences in accuracy. 

For the power control case, we found subdividing the outputs by region (to lower the output dimensionality of an otherwise 22 output problem) largely corrected the gap in performance between KAN and FNN for this dataset, suggesting that KANs may be weaker for problems with relatively few inputs compared to outputs or perhaps just for very high output problems. We show the specifics of these results in \ref{sec:metrics}. Future work should seek to clarify the cause and specific nature of this shortcoming and whether there are any rules of thumb that KAN follows for input and output dimensionality to achieve optimal performance.

For the case of the light water reactor, the size of the dataset may have inhibited both the FNN and the KAN to learn accurate prediction mechanisms for the Max-Fxy output. This combined with comparable (high) performance across remaining datasets for all outputs supports the claim that KANs are comparable in their prediction accuracy to FNNs, with a notable exception to problems with many outputs.   

\subsection{Interpretability}
In Section \ref{sec:results}, we describe the model metrics and explainability analysis for the KAN and the FNN models of the heat conduction dataset, as well as the numerical analysis method used to derive the heat conduction dataset values and the equation created by the KAN. Though we found it difficult to see a direct link between the equations for the numerical analysis and the equation derived by KAN, the KAN method of determining $T$, the temperature in the nuclear fuel rod, was obviously more straightforward. The SHAP explainability analysis for the heat conduction dataset highlights this simplicity, especially compared to the FNN method of prediction. The KAN actually eliminated three of the seven variables used to generate the dataset, and then predicted the values with an $R^2$ score of 0.996, surpassing the FNN. This reduction in dimension and ability to plainly yield an equation that so accurately predicts an output exemplifies the interpretability of the KAN. A human could read and reproduce the results of the KAN model in this work by plugging in values into Eq.\ref{eq:HEAT_KAN}. No other machine learning method offers this level of insight into the inner-workings of the model while still offering high accuracy. Yet, this level of insight precisely enables regulatory agencies, engineers, doctors, and scientists alike to truly understand, manage, and trust such a model. As a caveat, future work should inquire how to make equations like Eq.\eqref{eq:HEAT_KAN} \textit{more} interpretable for specific audiences. Flashing a lengthy equation at a patient does not meet the comprehensibility requirements we described in Section \ref{sec:intro}, but it does provide a good place to start.

A quick look into \ref{app:equations} makes it clear that not all KAN equations are created equal. Despite pruning and regularization, operations that sparsify (i.e., simplify) a KAN network, many of these equations still extend many lines, as we saw in Eq.\eqref{eq:f-delta-h_BWR} for the light water reactor dataset. This naturally raises the question: is a long equation still an interpretable equation? Based on the definition we outline in Section \ref{sec:theory}, technically, yes. However, as an equation grows in length, it loses value in computational feasibility. Further, an upper limit to interpretability surely exists as equation length tends to infinity, the point where a human can no longer reproduce the result in a reasonable time constraint. Fortunately, we can print all the equations in this work on a finite amount of paper, but future work should be aware that deeper KANs produce lengthier equations. Those seeking to favor simplicity over accuracy have options in both KAN hyperparameter tuning and architecture development to achieve such objectives. 

\subsection{Explainability}
In Section \ref{sec:results}, we describe the model metrics and explainability analysis for the KAN and the FNN models of the CHF dataset. Both models perform extremely well for CHF, with some larger absolute errors from the KAN. As the only experimentally gathered dataset in this work, the CHF results provide a unique opportunity to evaluate the physicality of both KAN and FNN models, particularly since no known closed form solution exists for the prediction of CHF, but correlations, such as the W-3 correlation presented in Section \ref{sec:results} do. Figure \ref{fig:chf_explain} shows the SHAP analysis of the KAN and FNN, respectively. Myers et al. \cite{myers_pymaise_2025} performed a similar analysis of an FNN trained with the same hyperparameters on this dataset, but varied the set of input features to exclude either $T\_in$ or $Xe$, as shown in Figures 7 and 8 of that work \cite{myers_pymaise_2025}. They found that, for an FNN, eliminating $T\_in$ increased the importance of both $Xe$ and $P$ and eliminating Xe had almost no effect on the importance of $T\_in$ and $P$ when compared to the rankings of the full set of input variables. In Figure \ref{fig:chf_explain}, which represents the full dataset, we observe for the FNN results that $T\_in$ ranks higher than $Xe$, which ranks higher than $P$. However, from both the variational study performed by Myers et al. \cite{myers_pymaise_2025} and thermodynamic knowledge, we know that inlet temperature, outlet quality, and pressure are correlated values. Yet, the FNN fails to emphasize the importance of pressure in predicting CHF. Taken at face value, the FNN's explainability metrics would suggest an insignificant distinction between the safety calculation of a pressurized water reactor and a boiling water reactor, an obviously false assumption.

While on one hand this highlights a known limitation of post-hoc explainability analysis for FNNs, the KAN results shown in Figure \ref{fig:chf_explain} tell a different story, demonstrating heightened importance for five of the six input features, even when they are correlated. By using SHAP as an explainability tool for both the FNN and the KAN, we simultaneously use it as an over-engineered sensitivity tool for the equation produced by KAN, an equation that encompasses physical relations. Though it still produces impressive results, the FNN merely yields statistical accuracy. Liu et al. \cite{liu_kan_2024} demonstrate KAN's ability to re-derive known physical relationships; this work extends that feature beyond a toy problem into a real one with no known solution.  

As mentioned previously, future work should consider alternative explainability methods that more accurately reflect classical SHAP values and/or methods that capture deeper explainability metrics, like node and edge attribution scores.

\subsection{Practical Implications}

What advantages do KAN’s symbolic equations offer to practical applications that traditional black-box ML models, including FNNs, do not? One key benefit is the ability to compute derivatives directly, thanks to the closed-form nature of symbolic equations. This is particularly valuable for real-time applications requiring exact gradients, which are often difficult to obtain from other black-box models. While this advantage is somewhat constrained by the length and complexity of the equations, most functions used in KAN’s equations are well-known with readily available derivatives.

The notable aspect of KAN’s explainability is its reproducibility—even without methods like SHAP, which we applied here for consistency with FNN, it remains straightforward to isolate the contribution of each feature or input. Unlike traditional black-box models that require advanced post-hoc methods, KAN allows direct identification of feature impacts by isolating terms in its equations. More importantly, since KAN equations are not dependent on random seeding, users can expect consistent results, leading to deterministic feature importance assessments.

Perhaps the most valuable application of KAN’s symbolic features in the energy sector lies in developing more advanced correlation techniques that can uncover hidden patterns in data more effectively than traditional least squares or polynomial fitting. Eq.\eqref{eq:CHF} for CHF still resembles a semi-empirical formula, similar to the W-3 correlation, except that KAN is expected to provide better interpolation and capture non-linearity more effectively. Therefore, it is fair to say that KAN does not replace the underlying physics, mathematical models, or computational codes but rather transforms how correlations are constructed.

The light water reactor dataset and the complex KAN equation in Eq.\eqref{eq:f-delta-h_BWR} demonstrate that KAN equations can become quite intricate and computationally expensive to evaluate. However, this complexity is not surprising given the underlying physics being modeled. As previously discussed, estimating F-delta-H might appear to be a straightforward ratio in Eq.\eqref{eq:fdeltah_physics}, but the physics behind it is inherently complex and computationally demanding. Consequently, it is unrealistic to expect KAN to produce a significantly simpler representation. Typically, using nuclear codes for this estimation can take 20–30 minutes (excluding modeling time) to generate an accurate result. In this context, KAN remains a faster alternative for computation.

A potential criticism of this work is its focus on nuclear-related datasets. While we have argued that these datasets are multidisciplinary, non-redundant, and present a diverse set of challenges typical of machine learning problems, this justification may need further elaboration. The primary reason for this focus is that the authors prioritized establishing meaningful physical connections with KAN results—an area where our team has confidence due to our nuclear expertise—rather than simply demonstrating KAN’s performance on randomly selected but diverse datasets. While we could have taken that approach, we believe it would offer less value. We encourage researchers in other sensitive industries and renewable energy sectors to extend this work to their domains, where they can confidently interpret and justify the results. Potential renewable energy applications may include solar systems \cite{rong2025recurrent}, wind energy \cite{mubarak2024quasi}, hydrogen energy \cite{ma2018data}, and hybrid energy systems \cite{radaideh2020design}.

Another limitation to note is that we did not observe either KAN or FNN extracting more information beyond interpolating between the provided dataset points. This means we cannot guarantee extrapolation performance—a fundamental limitation of all machine learning models. While developing Bayesian methods to support KAN’s uncertainty quantification and assess interpolation error, similar to Gaussian Processes \cite{rasmussen2003gaussian,radaideh2020surrogate}, could provide valuable insights, it would not necessarily eliminate the extrapolation weakness. Moreover, such an approach would likely increase computational costs, data requirements, and the complexity of KAN’s formulation, much like what occurs with Bayesian neural networks \cite{blundell2015weight}.
\section{Conclusion}
\label{sec:conclusion}
Sensitive industries, like medicine, finance, energy, and engineering demand machine learning models that allow them to keep up with industry without sacrificing their integrity. This work proposes a KAN to meet that demand for its interpretability and explainability when coupled with post-hoc methods, like SHAP. After testing a KAN against an FNN on eight nuclear engineering datasets as sensitive case studies that also offer multidisciplinary aspects, we found both models comparable in terms of accuracy when some adjustments are made for limiting the output space size. We also found KANs better at capturing known physical relationships than FNNs, which instead feature mere statistical accuracy. We unexpectedly observed that KANs may also enhance interpretability of problems with known solutions by generating models with reduced dimensions, via the elimination of features in equation generation. 

In addition to the direct calls for future work throughout this paper, future studies should consider testing KANs against additional models and additional datasets. Additional models should include physics-informed neural networks, which may better handle physical relationships than a simple FNN, and additional datasets should consider more experimental datasets, not only from engineering, but from medical and financial applications as well. Further, though cautioned by the literature, this analysis found no issues with time complexity or excessive computational resource demand. However, this could be due to sufficiently small datasets. Future work might test this with larger datasets to more completely assess these impacts.

Overall, we find KANs a promising, interpretable alternative to traditional machine learning methods for sensitive applications of a multidisciplinary nature.

\section*{Data Availability}
\label{sec:availability}

Currently, the authors possess, in a private GitHub repository, all the data and codes needed to reproduce all the results in this work. To ensure confidentiality of this research, the authors will make this repository public during an advanced stage of the review process, and it will be listed under our research group's public Github page: \url{https://github.com/aims-umich}.

\section*{Acknowledgment}

This research is funded by the U.S. Nuclear Regulatory Commission's University Nuclear Leadership Program for Research and Development, award number 31310024M0013. The first author (N. Panczyk) received sponsorship through the National Science Foundation's Graduate Research Fellowship Program.

\section*{CRediT Author Statement}

\begin{itemize}
    \item \textbf{Nataly R. Panczyk}: Conceptualization, Methodology, Software, Validation, Formal Analysis, Visualization, Investigation, Data Curation, Writing - Original Draft. 
    \item \textbf{Omer F. Erdem}: Methodology, Software, Validation, Formal Analysis, Investigation, Writing - Review and Edit.  
    \item \textbf{Majdi I. Radaideh}: Conceptualization, Methodology, Data Curation, Formal Analysis, Resources, Funding Acquisition, Supervision, Project Administration, Writing - Review and Edit. 
\end{itemize}

\bibliographystyle{elsarticle-num}
\setlength{\bibsep}{0pt plus 0.2ex}
{
\bibliography{references}}


\appendix
\section{Performance Metrics for Other Datasets}
\label{sec:metrics}

\vspace{-10mm}

\begin{longtable}{llrrrrrr}
\label{tab:fp_metrics} \\
\caption{Performance metrics for the \textbf{Materials and Fuel Performance} dataset as modeled by both a \textbf{symbolic} KAN and FNN.} \\
\toprule
Output & Model & MAE & MAPE & MSE & RMSE & RMSPE & $R^2$ \\
\midrule
\endfirsthead
\toprule
Output & Model & MAE & MAPE & MSE & RMSE & RMSPE & $R^2$ \\
\midrule
\endhead
\midrule
\multicolumn{8}{r}{Continued on next page} \\
\midrule
\endfoot
\bottomrule
\endlastfoot
fission\_gas & KAN & 5.33434e-08 & 0.17391 & 6.63868e-15 & 8.14781e-08 & 0.26956 & 0.99729 \\
fission\_gas & FNN & 8.76018e-08 & 0.28551 & 1.43874e-14 & 1.19947e-07 & 0.39009 & 0.99413 \\
\midrule
max\_fuel\_cl\_T & KAN & 1.89257 & 0.11954 & 7.55531 & 2.74869 & 0.17633 & 0.99377 \\
max\_fuel\_cl\_T & FNN & 1.90921 & 0.11984 & 6.17674 & 2.48530 & 0.15621 & 0.99491 \\
\midrule
max\_fuel\_surf\_T & KAN & 0.89289 & 0.12679 & 1.59051 & 1.26115 & 0.17867 & 0.97612 \\
max\_fuel\_surf\_T & FNN & 1.48399 & 0.21102 & 3.86673 & 1.96640 & 0.27942 & 0.94195 \\
\midrule
radial\_clad\_T & KAN & 4.97319e-08 & 0.26086 & 5.46257e-15 & 7.39092e-08 & 0.40105 & 0.99061 \\
radial\_clad\_T & FNN & 3.73859e-08 & 0.19566 & 2.17464e-15 & 4.66329e-08 & 0.24597 & 0.99626 \\
\end{longtable}

\vspace{-10mm}

\begin{longtable}{llrrrrrr}
\label{tab:htgr_metrics} \\
\caption{Performance metrics for the \textbf{Microreactor} dataset as modeled by both a \textbf{symbolic} KAN and FNN.} \\
\toprule
Output & Model & MAE & MAPE & MSE & RMSE & RMSPE & $R^2$ \\
\midrule
\endfirsthead
\toprule
Output & Model & MAE & MAPE & MSE & RMSE & RMSPE & $R^2$ \\
\midrule
\endhead
\midrule
\multicolumn{8}{r}{Continued on next page} \\
\midrule
\endfoot
\bottomrule
\endlastfoot
fluxQ1 & KAN & 0.000337 & 0.134820 & 1.76464e-07 & 0.000420 & 0.167857 & 0.989950 \\
fluxQ1 & FNN & 0.000510 & 0.203661 & 4.29808e-07 & 0.000656 & 0.261368 & 0.975521 \\
\midrule
fluxQ2 & KAN & 0.000360 & 0.143924 & 2.04094e-07 & 0.000452 & 0.180149 & 0.988376 \\
fluxQ2 & FNN & 0.000505 & 0.201829 & 4.17823e-07 & 0.000646 & 0.257740 & 0.976204 \\
\midrule
fluxQ3 & KAN & 0.000352 & 0.140688 & 1.99838e-07 & 0.000447 & 0.178268 & 0.988619 \\
fluxQ3 & FNN & 0.000495 & 0.197698 & 3.97459e-07 & 0.000630 & 0.251424 & 0.977364 \\
\midrule
fluxQ4 & KAN & 0.000348 & 0.139110 & 1.99082e-07 & 0.000446 & 0.177860 & 0.988662 \\
fluxQ4 & FNN & 0.000524 & 0.209305 & 4.49469e-07 & 0.000670 & 0.267523 & 0.974401 \\
\end{longtable}

\vspace{-10mm}

\begin{longtable}{llrrrrrr}
\label{tab:mitr_A_metrics} \\
\caption{Performance metrics for the \textbf{Power Control dataset (Region A)} in Figure \ref{fig:MITR_schematic} as modeled by both a \textbf{symbolic} KAN and FNN.} \\
\toprule
Output & Model & MAE & MAPE & MSE & RMSE & RMSPE & $R^2$ \\
\midrule
\endfirsthead
\toprule
Output & Model & MAE & MAPE & MSE & RMSE & RMSPE & $R^2$ \\
\midrule
\endhead
\midrule
\multicolumn{8}{r}{Continued on next page} \\
\midrule
\endfoot
\bottomrule
\endlastfoot
A-2 & KAN & 3.42780 & 0.01330 & 21.59370 & 4.64690 & 0.01800 & 0.99800 \\
A-2 & FNN & 3.32124 & 0.01285 & 18.30655 & 4.27862 & 0.01657 & 0.99829 \\
\end{longtable}

\vspace{-10mm}

\begin{longtable}{llrrrrrr}
\label{tab:mitr_B_metrics} \\
\caption{Performance metrics for the \textbf{Power Control dataset (Region B)} in Figure \ref{fig:MITR_schematic} as modeled by both a \textbf{symbolic} KAN and FNN. \textit{Four selected outputs are reported for brevity.}} \\
\toprule
Output & Model & MAE & MAPE & MSE & RMSE & RMSPE & $R^2$ \\
\midrule
\endfirsthead
\toprule
Output & Model & MAE & MAPE & MSE & RMSE & RMSPE & $R^2$ \\
\midrule
\endhead
\midrule
\multicolumn{8}{r}{Continued on next page} \\
\midrule
\endfoot
\bottomrule
\endlastfoot
B-1 & KAN & 5.21890 & 0.02300 & 54.29580 & 7.36860 & 0.03250 & 0.99710 \\
B-1 & FNN & 5.67038 & 0.02491 & 49.71682 & 7.05102 & 0.03097 & 0.99738 \\
\midrule
B-2 & KAN & 4.64950 & 0.02160 & 37.90760 & 6.15690 & 0.02860 & 0.99820 \\
B-2 & FNN & 5.26452 & 0.02441 & 43.51394 & 6.59651 & 0.03060 & 0.99794 \\
\midrule
B-4 & KAN & 4.79840 & 0.02100 & 39.52180 & 6.28660 & 0.02760 & 0.99730 \\
B-4 & FNN & 5.09374 & 0.02233 & 40.32098 & 6.34988 & 0.02786 & 0.99723 \\
\midrule
B-5 & KAN & 4.17800 & 0.01910 & 32.83440 & 5.73010 & 0.02620 & 0.99810 \\
B-5 & FNN & 3.99387 & 0.01828 & 26.70497 & 5.16769 & 0.02363 & 0.99846 \\
\end{longtable}

\vspace{-10mm}

\begin{longtable}{llrrrrrr}
\label{tab:mitr_C_metrics} \\
\caption{Performance metrics for the \textbf{Power Control dataset (Region C)} in Figure \ref{fig:MITR_schematic} as modeled by both a \textbf{symbolic} KAN and FNN. \textit{Four selected outputs are reported for brevity.}} \\
\toprule
Output & Model & MAE & MAPE & MSE & RMSE & RMSPE & $R^2$ \\
\midrule
\endfirsthead
\toprule
Output & Model & MAE & MAPE & MSE & RMSE & RMSPE & $R^2$ \\
\midrule
\endhead
\midrule
\multicolumn{8}{r}{Continued on next page} \\
\midrule
\endfoot
\bottomrule
\endlastfoot
C-1 & KAN & 45.15530 & 0.25130 & 3107.75010 & 55.74720 & 0.31100 & 0.96510 \\
C-1 & FNN & 6.90512 & 0.03844 & 87.83895 & 9.37224 & 0.05242 & 0.99901 \\
\midrule
C-2 & KAN & 26.25600 & 0.14500 & 1063.05380 & 32.60450 & 0.17980 & 0.99160 \\
C-2 & FNN & 8.23736 & 0.04563 & 114.61896 & 10.70602 & 0.05948 & 0.99909 \\
\midrule
C-3 & KAN & 26.46490 & 0.14170 & 1106.63130 & 33.26610 & 0.17810 & 0.99140 \\
C-3 & FNN & 11.71499 & 0.06292 & 208.80959 & 14.45024 & 0.07788 & 0.99838 \\
\midrule
C-4 & KAN & 26.53280 & 0.14190 & 1068.77430 & 32.69210 & 0.17450 & 0.99030 \\
C-4 & FNN & 12.73486 & 0.06816 & 232.57355 & 15.25036 & 0.08151 & 0.99789 \\
\end{longtable}

\vspace{-10mm}

\begin{longtable}{llrrrrrr}
\label{tab:rea_metrics} \\
\caption{Performance metrics for the \textbf{Nuclear Safety} dataset as modeled by both a \textbf{symbolic} KAN and FNN.} \\
\toprule
Output & Model & MAE & MAPE & MSE & RMSE & RMSPE & $R^2$ \\
\midrule
\endfirsthead
\toprule
Output & Model & MAE & MAPE & MSE & RMSE & RMSPE & $R^2$ \\
\midrule
\endhead
\midrule
\multicolumn{8}{r}{Continued on next page} \\
\midrule
\endfoot
\bottomrule
\endlastfoot
avg\_Tcool & KAN & 0.00890 & 0.00160 & 0.00020 & 0.01450 & 0.00260 & 0.99960 \\
avg\_Tcool & FNN & 0.01308 & 0.00233 & 0.00027 & 0.01647 & 0.00293 & 0.99951 \\
\midrule
burst\_width & KAN & 0.01420 & 4.20730 & 0.00040 & 0.01930 & 5.34890 & 0.98090 \\
burst\_width & FNN & 0.00620 & 1.46186 & 0.00053 & 0.02303 & 2.41617 & 0.97265 \\
\midrule
max\_TF & KAN & 1.57100 & 0.16800 & 4.74660 & 2.17870 & 0.22960 & 0.99630 \\
max\_TF & FNN & 0.50154 & 0.05626 & 2.54761 & 1.59612 & 0.20006 & 0.99803 \\
\midrule
max\_power & KAN & 6.02500 & 3.40670 & 82.42740 & 9.07900 & 6.67210 & 0.99810 \\
max\_power & FNN & 2.86917 & 2.02932 & 15.80721 & 3.97583 & 4.24141 & 0.99963 \\
\end{longtable}

\vspace{-10mm}

\begin{longtable}{llrrrrrr}
\label{tab:xs_metrics} \\
\caption{Performance metrics for the \textbf{Nuclear Cross-section} dataset as modeled by both a \textbf{symbolic} KAN and FNN.} \\
\toprule
Output & Model & MAE & MAPE & MSE & RMSE & RMSPE & $R^2$ \\
\midrule
\endfirsthead
\toprule
Output & Model & MAE & MAPE & MSE & RMSE & RMSPE & $R^2$ \\
\midrule
\endhead
\midrule
\multicolumn{8}{r}{Continued on next page} \\
\midrule
\endfoot
\bottomrule
\endlastfoot
k & KAN & 0.000056 & 0.004481 & 6.40897e-09 & 0.000080 & 0.006380 & 0.999889 \\
k & FNN & 0.000036 & 0.002894 & 1.26624e-08 & 0.000113 & 0.009037 & 0.999781 \\
\end{longtable}

\section{SHAP Plots for Other Datasets}
\label{app:shap-plots}

\begin{figure}[H]
\centering
\label{fig:htgr_explain}
\begin{subfigure}[b]{0.48\textwidth}
   \includegraphics[width=1\linewidth]{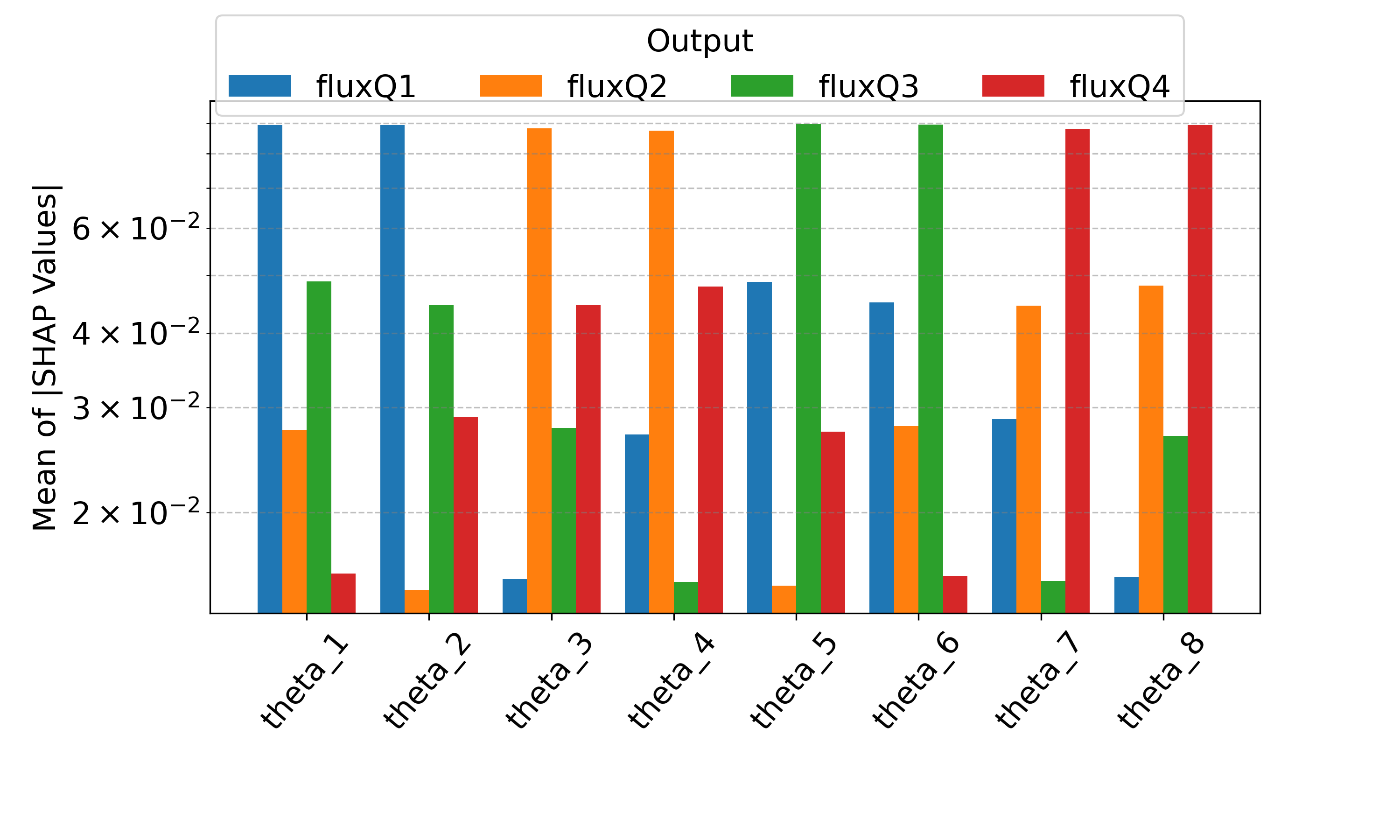}
   \caption{Symbolic KAN}
   \label{fig:htgr_kan} 
\end{subfigure}
\begin{subfigure}[b]{0.48\textwidth}
   \includegraphics[width=1\linewidth]{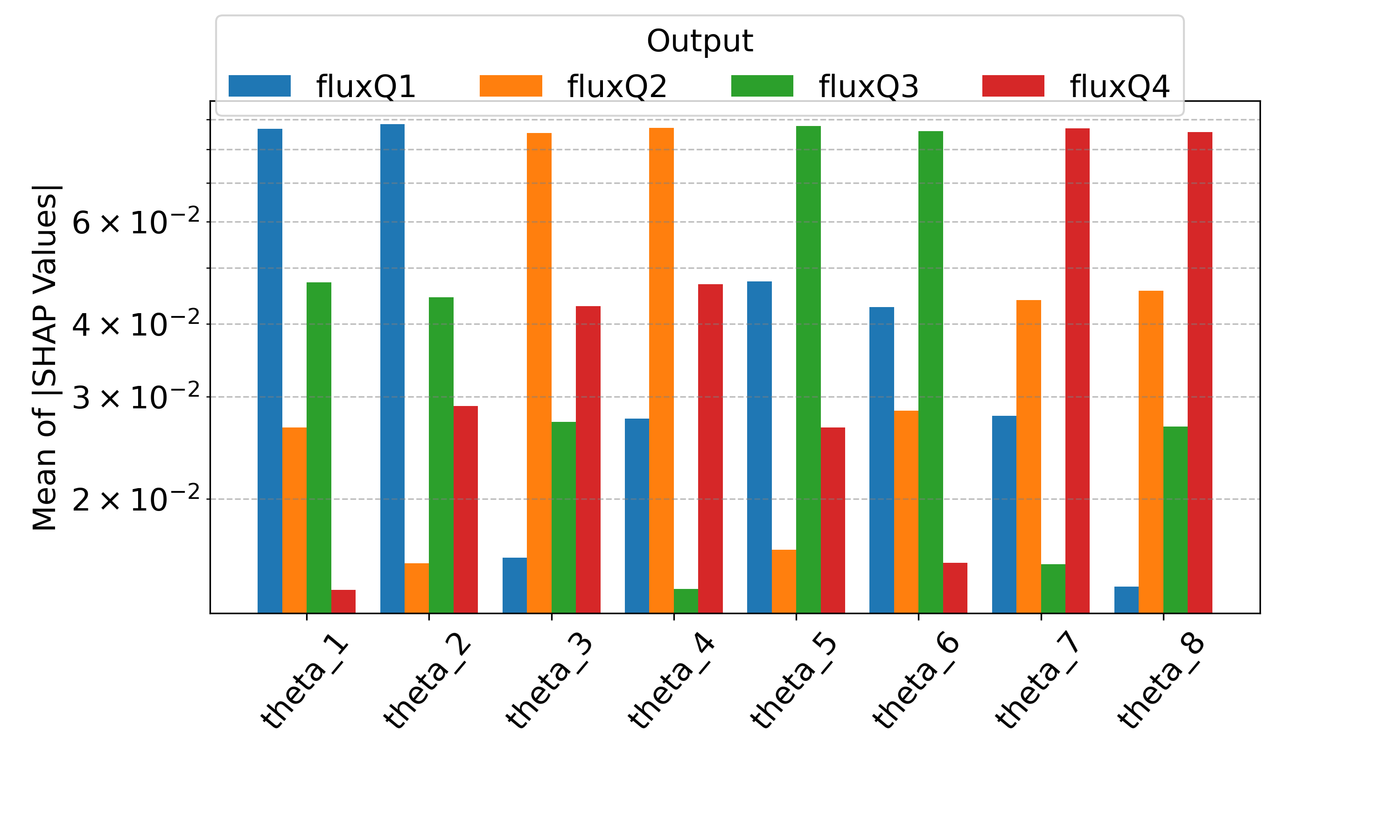}
   \caption{FNN}
   \label{fig:htgr_fnn}
\end{subfigure}
\caption[FP Explainability Results]{Feature importance ranking via SHAP for all outputs for both models of the \textbf{Microreactor} dataset.}
\end{figure}

\begin{figure}[H]
\centering
\label{fig:mitra_explain}
\begin{subfigure}[b]{0.48\textwidth}
   \includegraphics[width=1\linewidth]{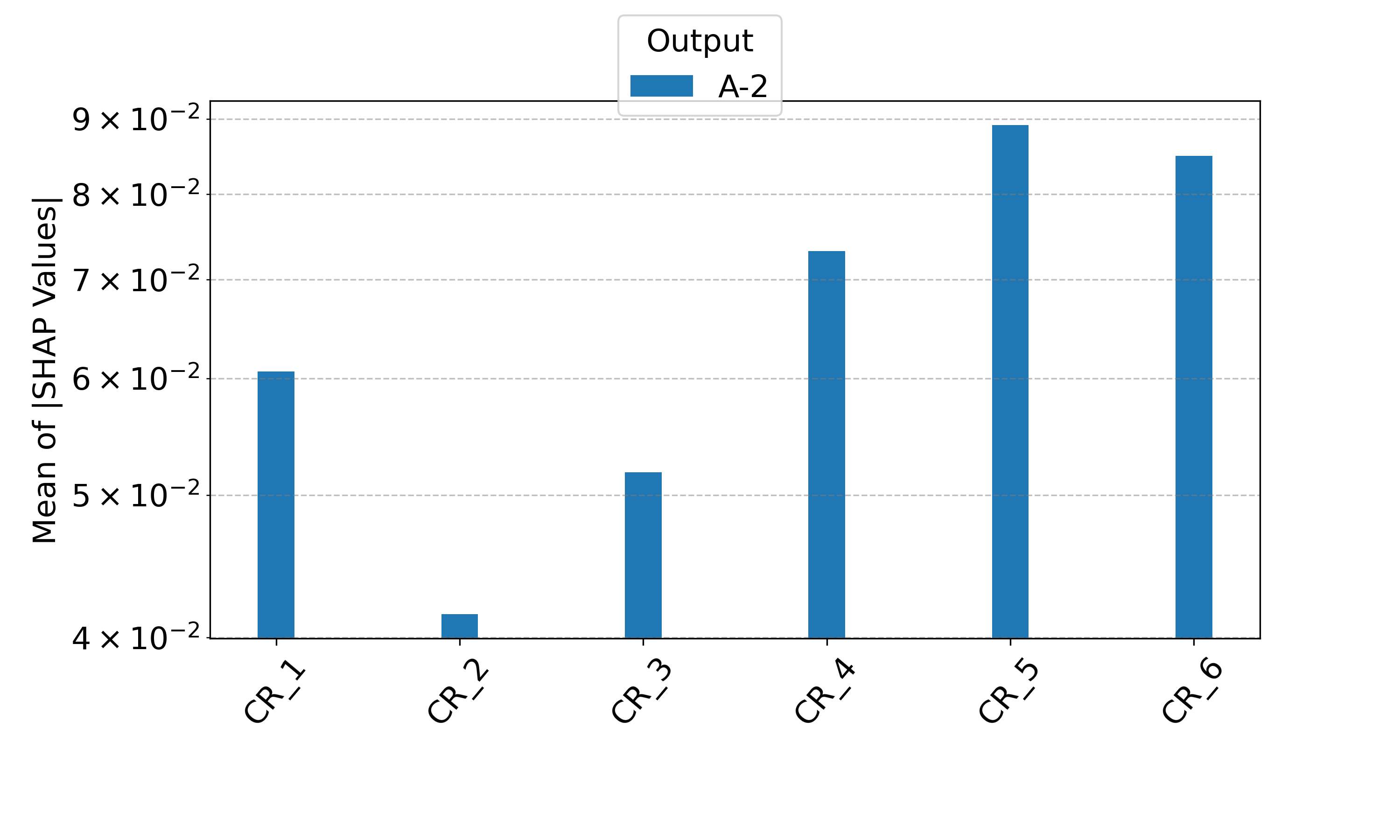}
   \caption{Symbolic KAN}
   \label{fig:mitra_kan} 
\end{subfigure}
\begin{subfigure}[b]{0.48\textwidth}
   \includegraphics[width=1\linewidth]{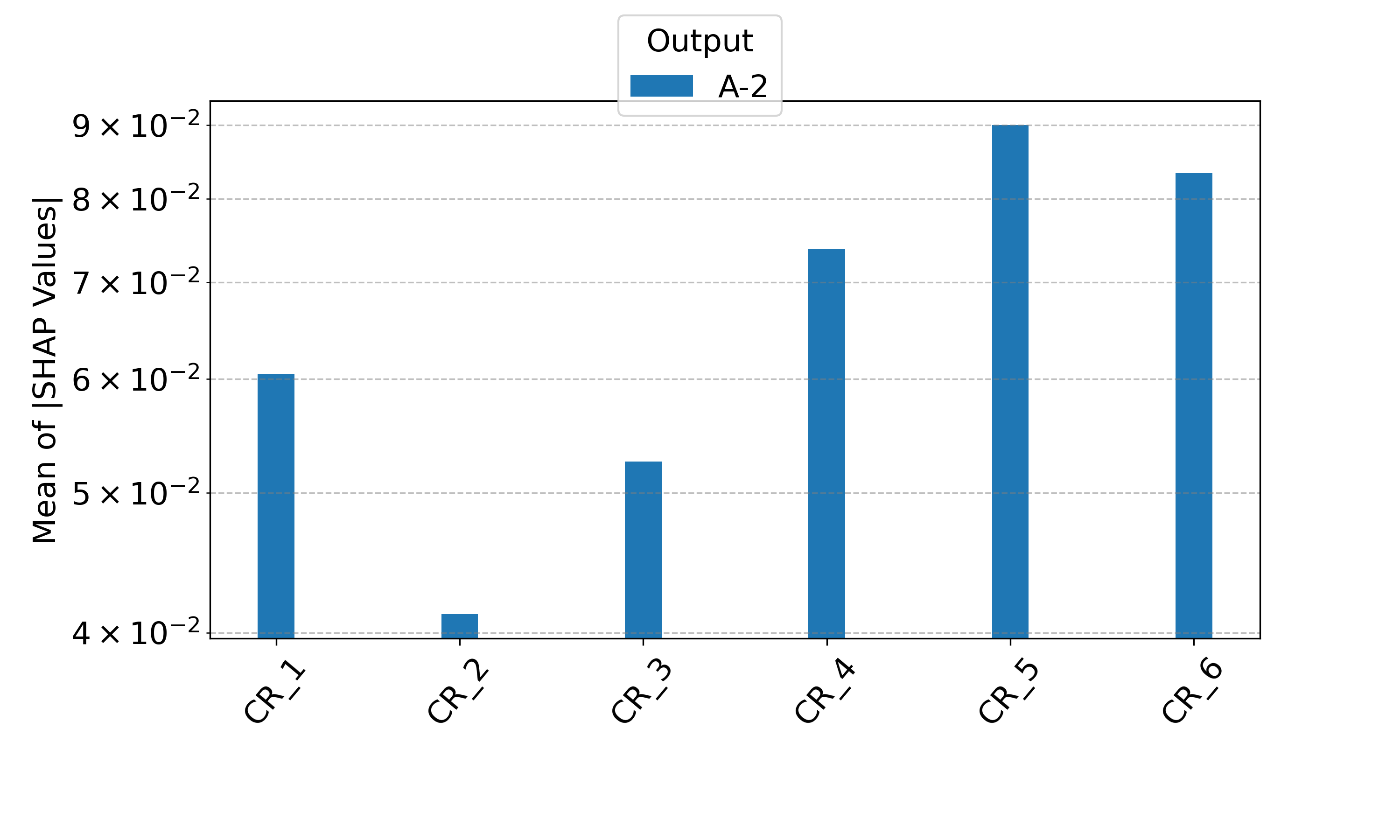}
   \caption{FNN}
   \label{fig:mitra_fnn}
\end{subfigure}
\caption[FP Explainability Results]{Feature importance ranking via SHAP for all outputs for both models of \textbf{Power Control dataset (Region A)}.}
\end{figure}

\begin{figure}[H]
\centering
\label{fig:mitrb_explain}
\begin{subfigure}[b]{0.48\textwidth}
   \includegraphics[width=1\linewidth]{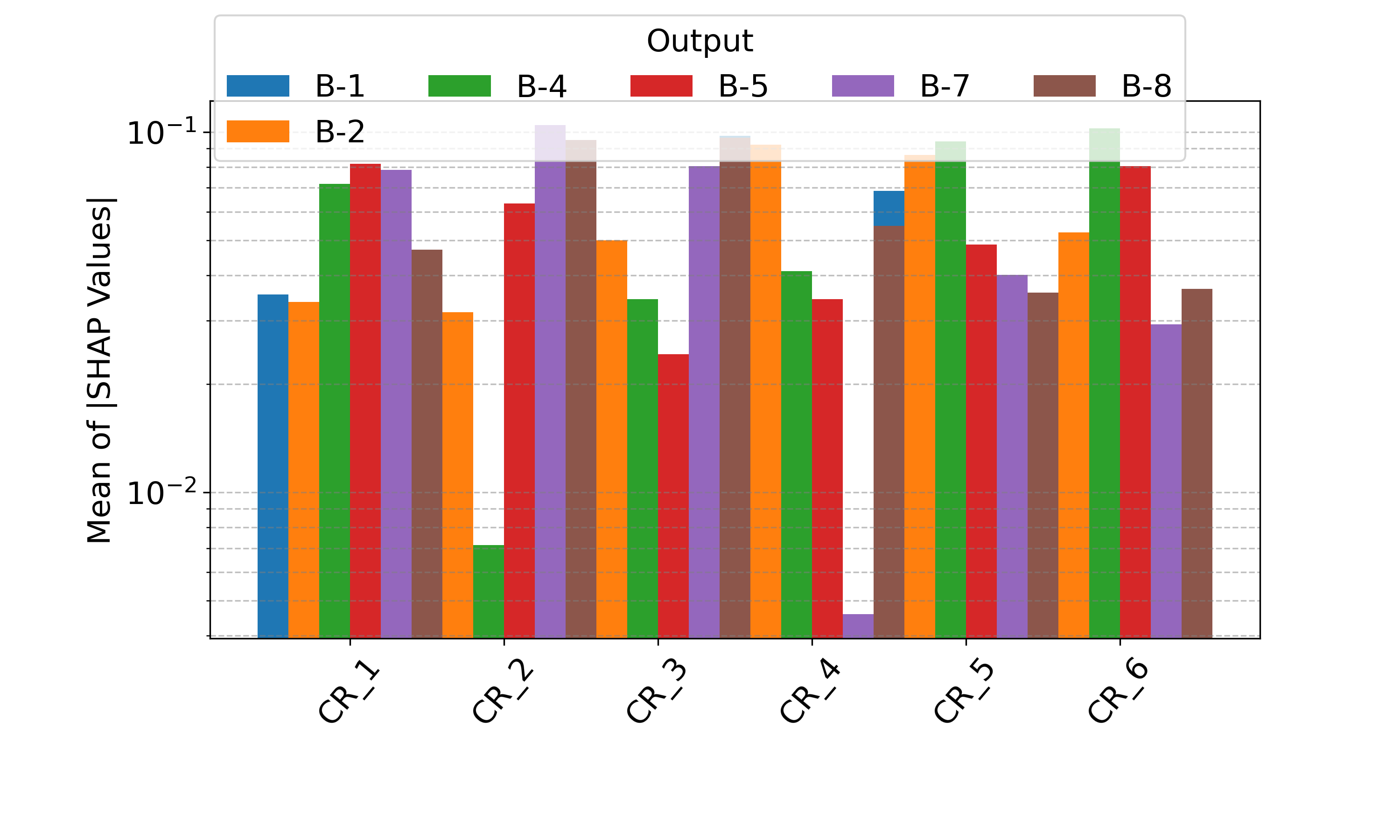}
   \caption{Symbolic KAN}
   \label{fig:mitrb_kan} 
\end{subfigure}
\begin{subfigure}[b]{0.48\textwidth}
   \includegraphics[width=1\linewidth]{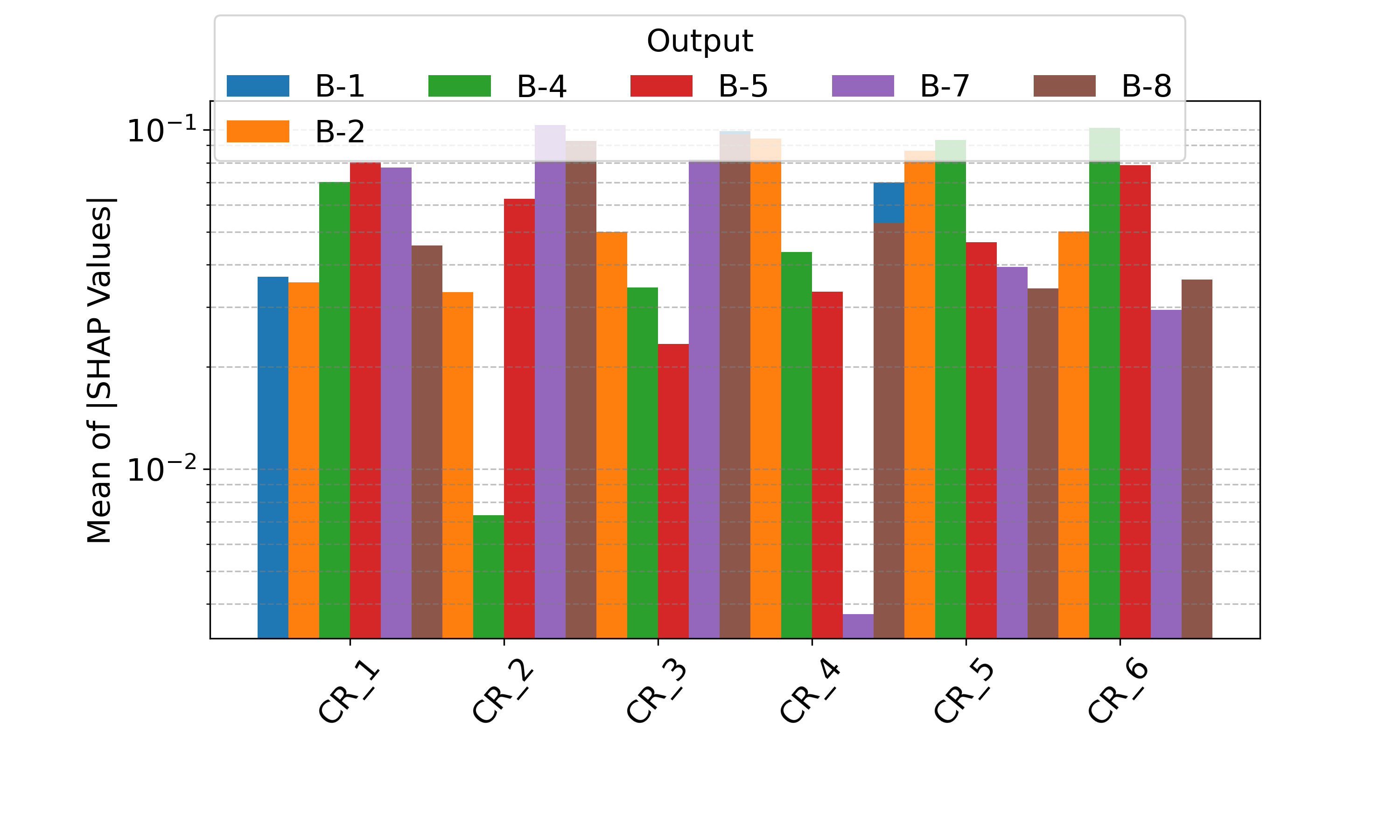}
   \caption{FNN}
   \label{fig:mitrb_fnn}
\end{subfigure}
\caption[FP Explainability Results]{Feature importance ranking via SHAP for all outputs for both models of the \textbf{Power Control dataset (Region B)}.}
\end{figure}

\begin{figure}[H]
\centering
\label{fig:mitrc_explain}
\begin{subfigure}[b]{0.48\textwidth}
   \includegraphics[width=1\linewidth]{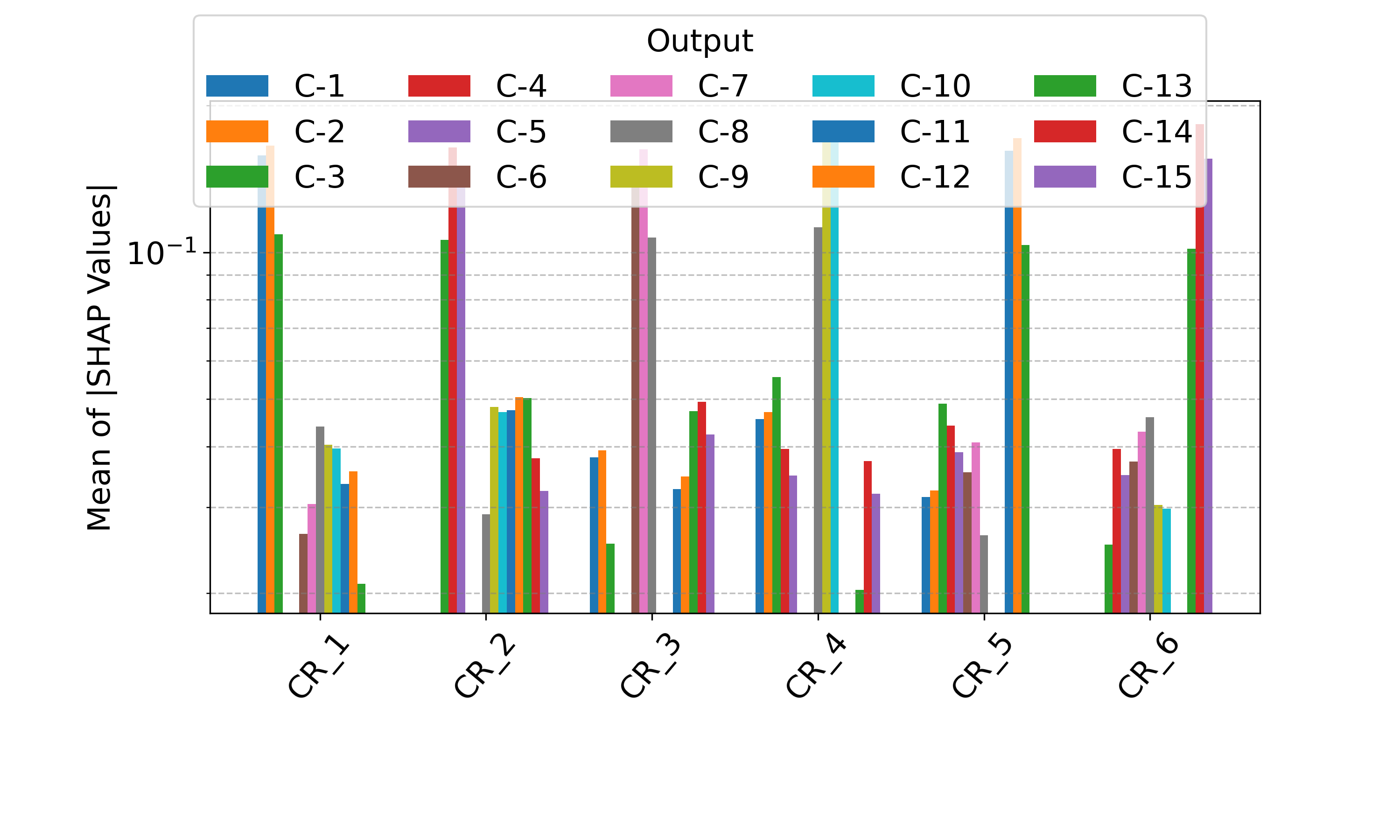}
   \caption{Symbolic KAN}
   \label{fig:mitrc_kan} 
\end{subfigure}
\begin{subfigure}[b]{0.48\textwidth}
   \includegraphics[width=1\linewidth]{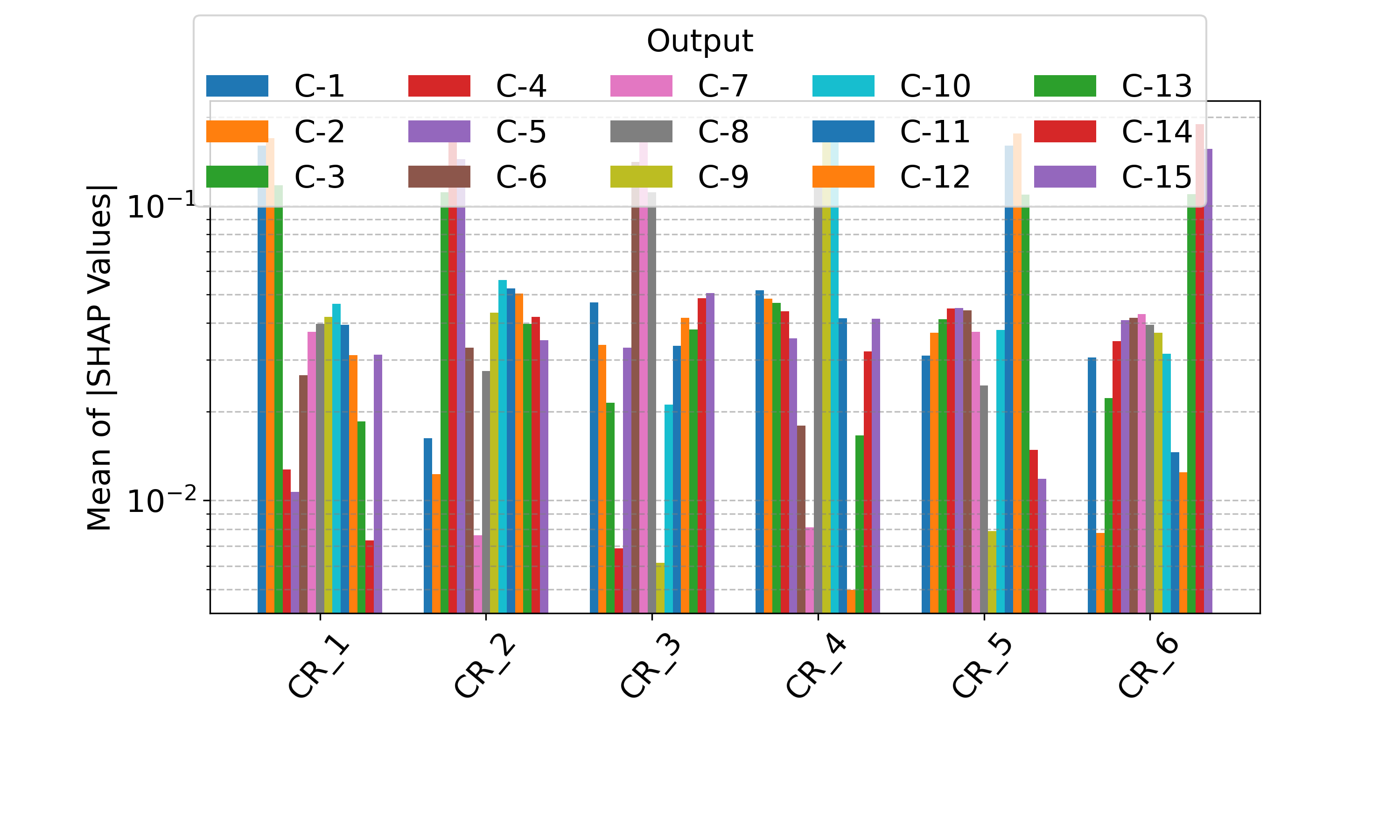}
   \caption{FNN}
   \label{fig:mitrc_fnn}
\end{subfigure}
\caption[FP Explainability Results]{Feature importance ranking via SHAP for all outputs for both models of the \textbf{Power Control dataset (Region C)}.}
\end{figure}

\begin{figure}[H]
\centering
\label{fig:rea_explain}
\begin{subfigure}[b]{0.48\textwidth}
   \includegraphics[width=1\linewidth]{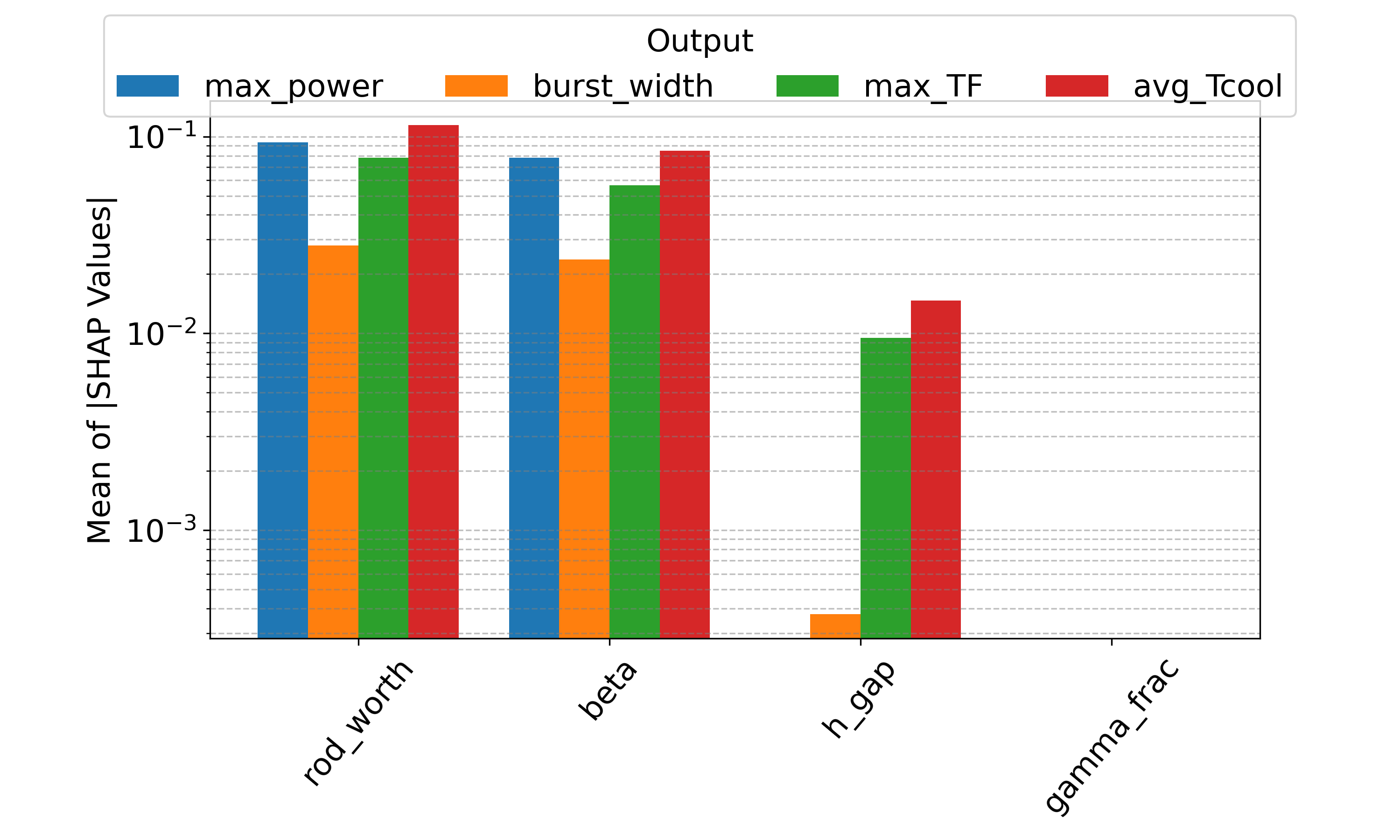}
   \caption{Symbolic KAN}
   \label{fig:rea_kan} 
\end{subfigure}
\begin{subfigure}[b]{0.48\textwidth}
   \includegraphics[width=1\linewidth]{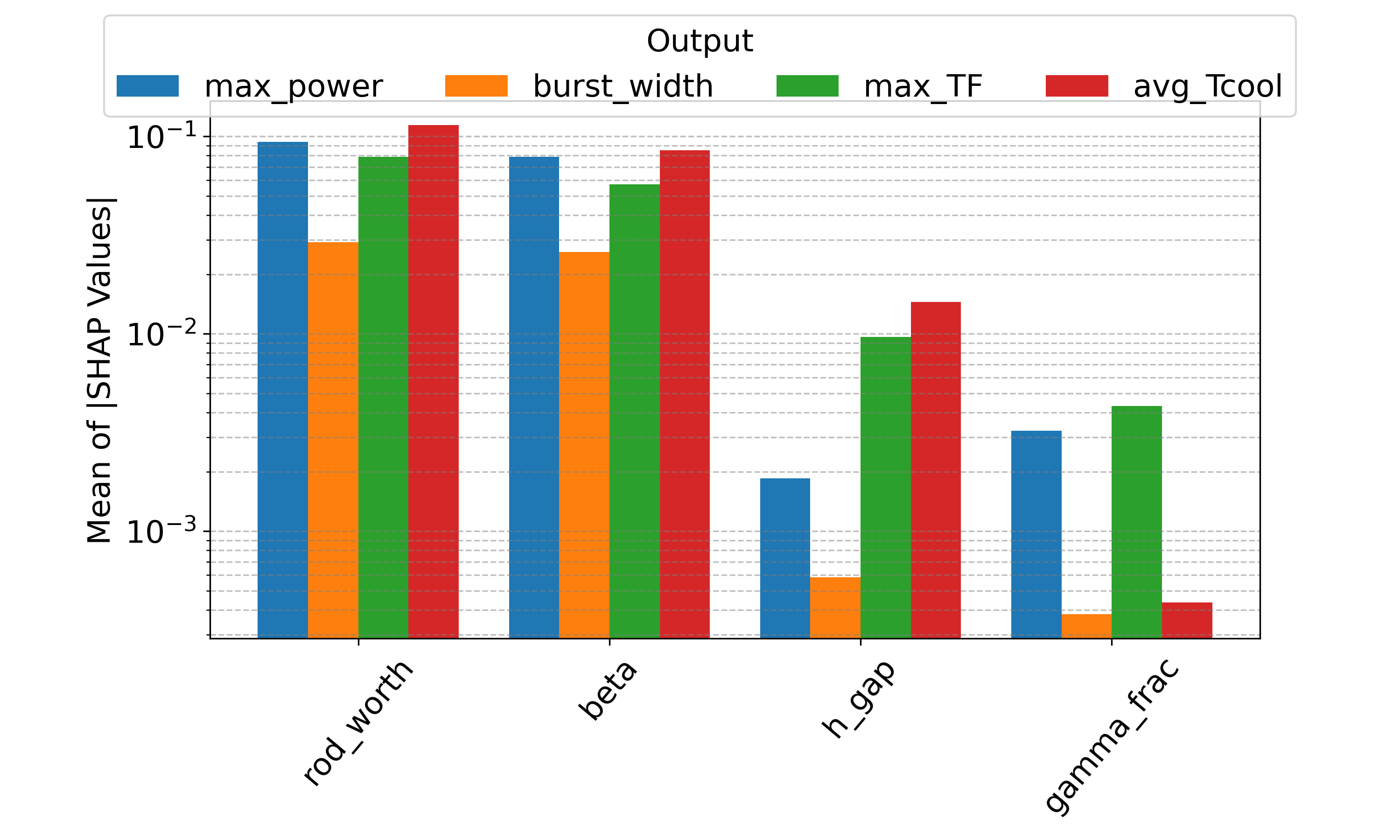}
   \caption{FNN}
   \label{fig:rea_fnn}
\end{subfigure}
\caption[FP Explainability Results]{Feature importance ranking via SHAP for all outputs for the both models of the \textbf{Nuclear Safety} dataset.}
\end{figure}

\begin{figure}[H]
\centering
\label{fig:xs_explain}
\begin{subfigure}[b]{0.48\textwidth}
   \includegraphics[width=1\linewidth]{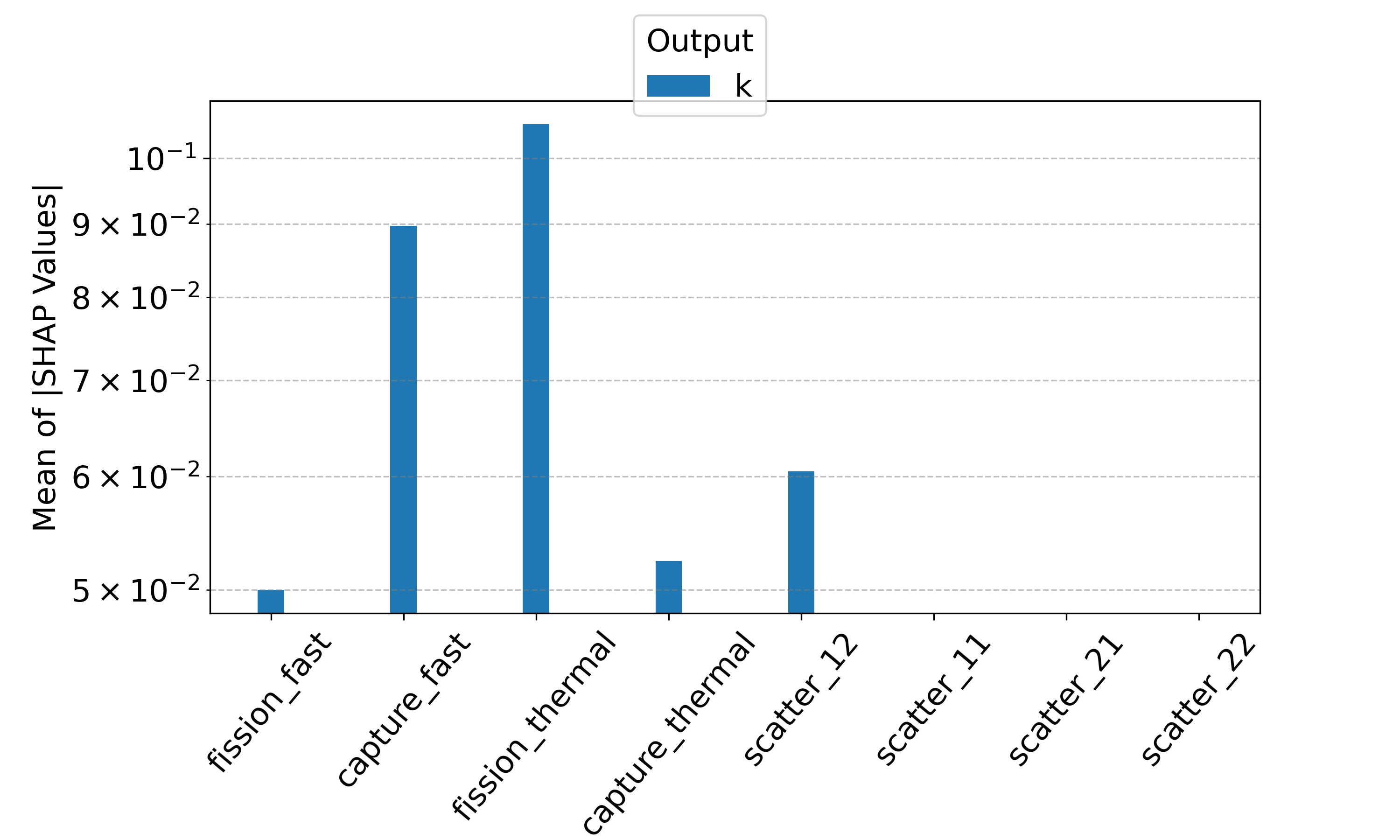}
   \caption{Symbolic KAN}
   \label{fig:xs_kan} 
\end{subfigure}
\begin{subfigure}[b]{0.48\textwidth}
   \includegraphics[width=1\linewidth]{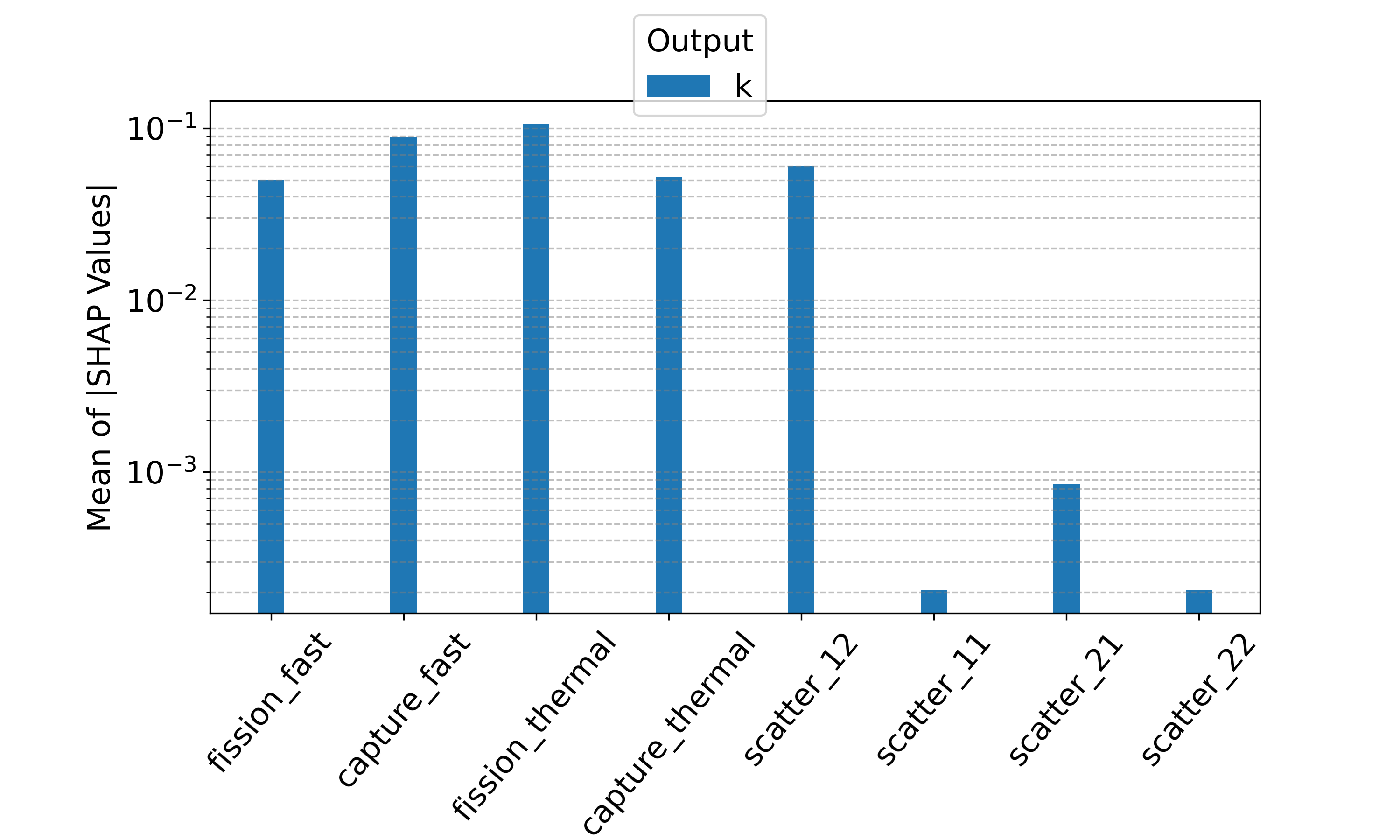}
   \caption{FNN}
   \label{fig:xs_fnn}
\end{subfigure}
\caption[FP Explainability Results]{Feature importance ranking via SHAP for all outputs for both models of the \textbf{Nuclear Cross-section} dataset.}
\end{figure}

\begin{figure}[H]
\centering
\label{fig:fp_explain}
\begin{subfigure}[b]{0.48\textwidth}
   \includegraphics[width=1\linewidth]{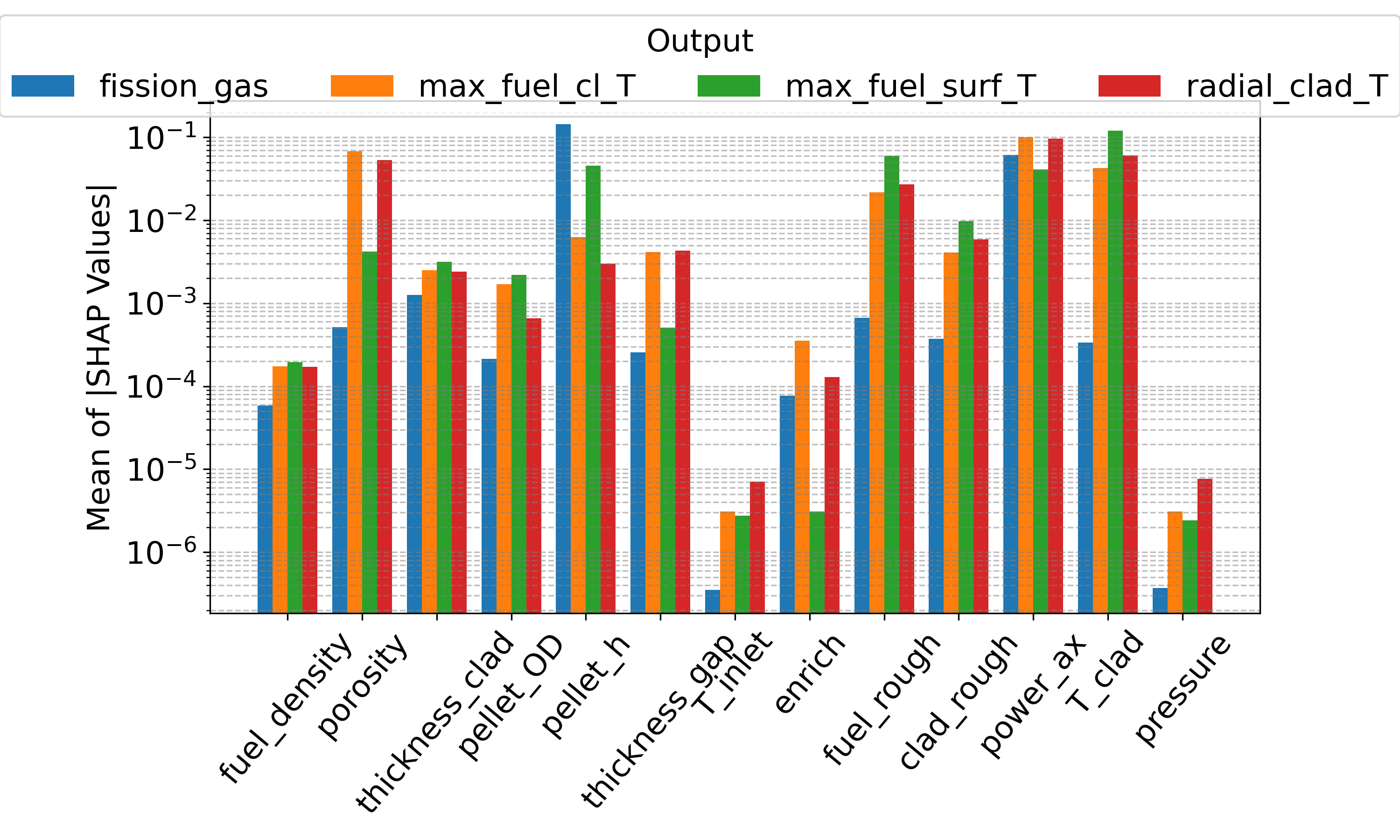}
   \caption{Symbolic KAN}
   \label{fig:fp_kan} 
\end{subfigure}
\begin{subfigure}[b]{0.48\textwidth}
   \includegraphics[width=1\linewidth]{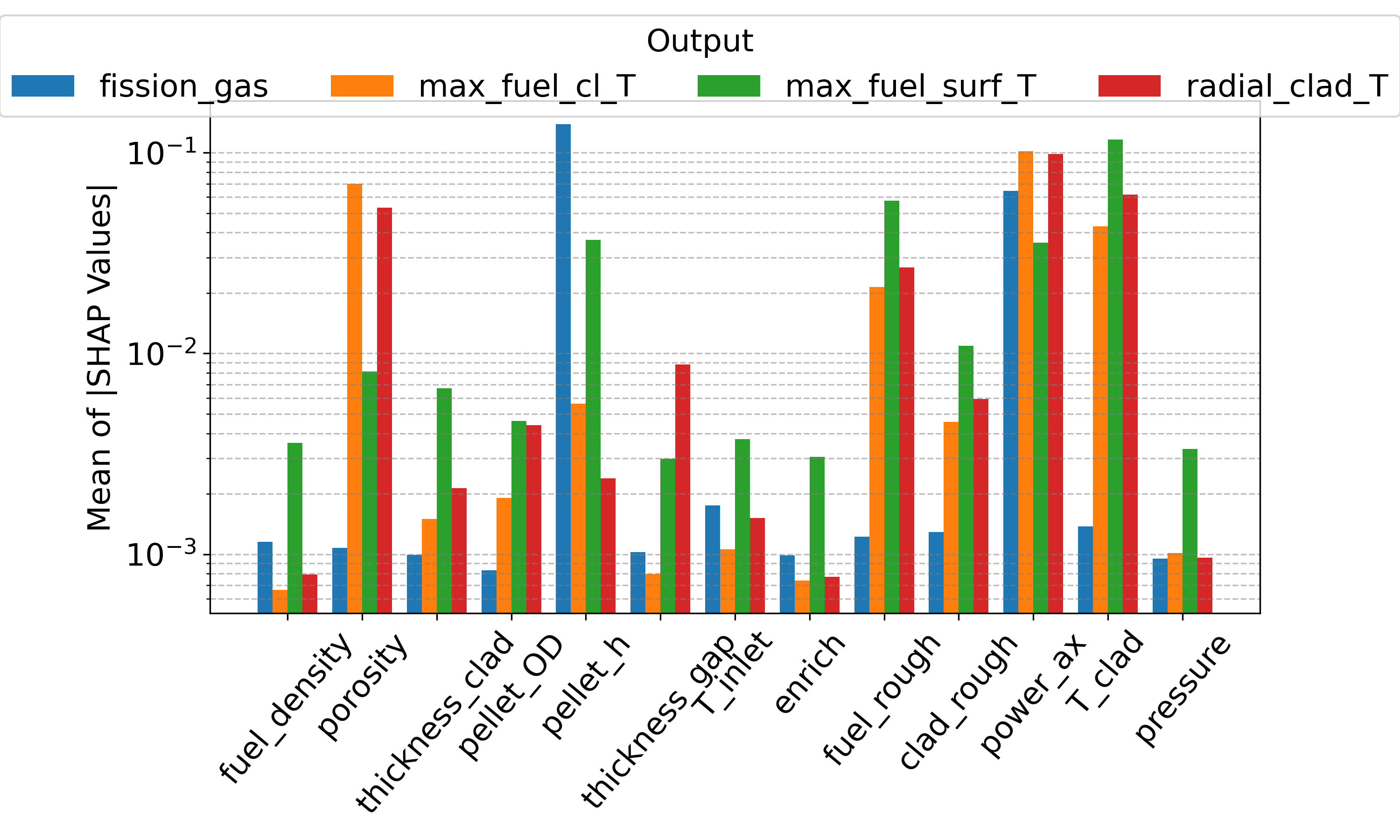}
   \caption{FNN}
   \label{fig:fp_fnn}
\end{subfigure}
\caption[FP Explainability Results]{Feature importance ranking via SHAP for all outputs for the both models of the \textbf{Materials and Fuel Performance} dataset.}
\end{figure}

\section{Symbolic Equations for Other Datasets}
\label{app:equations}

This appendix refers the reader to the Supplementary Material text file, "\verb|KAN_Equations.txt|", which contains additional examples of symbolic equations derived for other datasets that were not included in this work due to their extensive length. In addition to the supplementary text file, the complete set of equations generated in this research can be accessed through our GitHub repository, as detailed in the Data Availability section.

\end{document}